\documentclass[lettersize,journal]{IEEEtran}
\usepackage{amsmath,amsfonts}
\usepackage{algorithm}
\usepackage{array}
\usepackage[caption=false,font=normalsize,labelfont=sf,textfont=sf]{subfig}
\usepackage{textcomp}
\usepackage{stfloats}
\usepackage{url}
\usepackage{verbatim}
\usepackage{graphicx}
\usepackage{cite}
\newcommand{\etal}{et al.}
\newcommand{\ie}{i.e., }

\usepackage{bm} 
\usepackage{booktabs, multirow}
\usepackage{algpseudocode}
\makeatletter
\let\NAT@parse\undefined
\makeatother
\usepackage{hyperref}
\hypersetup{
    colorlinks=true,
    urlcolor=magenta,
}

\begin{document}
\bstctlcite{IEEEexample:BSTcontrol}

\title{Bidirectional Tutoring for Developmental Motor Learning in Robots: Co-Developed Interaction Dynamics Support Stable Learning}
\author{Rui Fukushima, Jun Tani
\thanks{R. Fukushima and J. Tani are with the Cognitive Neurorobotics Research Unit, Okinawa Institute of Science and Technology Graduate University, Okinawa, Japan.}
\thanks{Corresponding author: J. Tani. E-mail: jun.tani@oist.jp}
}



\maketitle
\begin{abstract}
Infants are well known to develop their motor skills through dense interaction with caregivers. Although such social interaction is crucial for human development, motor-skill learning in robots is often treated as a unidirectional process in which robots passively receive demonstrations from tutors. This overlooks a key property of social interaction: it is inherently bidirectional, with tutor and learner dynamically adapting to each other. In such interactions, the robot’s past experiences may function as prior constraints that shape the dynamics of their co-developed trajectories. We hypothesize that bidirectional tutoring allows such constraints to guide the formation of consistent behavioral patterns that preserve behavioral coherence and support generalization, whereas unidirectional interaction lacks such constraints and leads to broader, less consistent behavioral patterns. To examine this hypothesis, we conducted two experiments with a physical humanoid robot performing an object manipulation task: one involving human-robot interaction and another employing an AI tutor interacting with the real robot through an adaptive intervention mechanism designed to examine whether similar effects would emerge under more controlled conditions. We implement the developmental learning framework using a free-energy-principle-based neural network extended with generative replay, which supports stable sequence-by-sequence learning from single tutored episodes. Across both settings, bidirectional tutoring fostered consistent behaviors and stage-wise generalization, while the robot gradually required less tutor guidance. These results suggest that bidirectional tutoring, as an embodied and socially grounded approach, provides an effective scaffold for developmental motor learning in robots.
\end{abstract}

\section{Introduction}

\begin{figure}[ht] 
    \centering
    \includegraphics[width=0.5\textwidth]{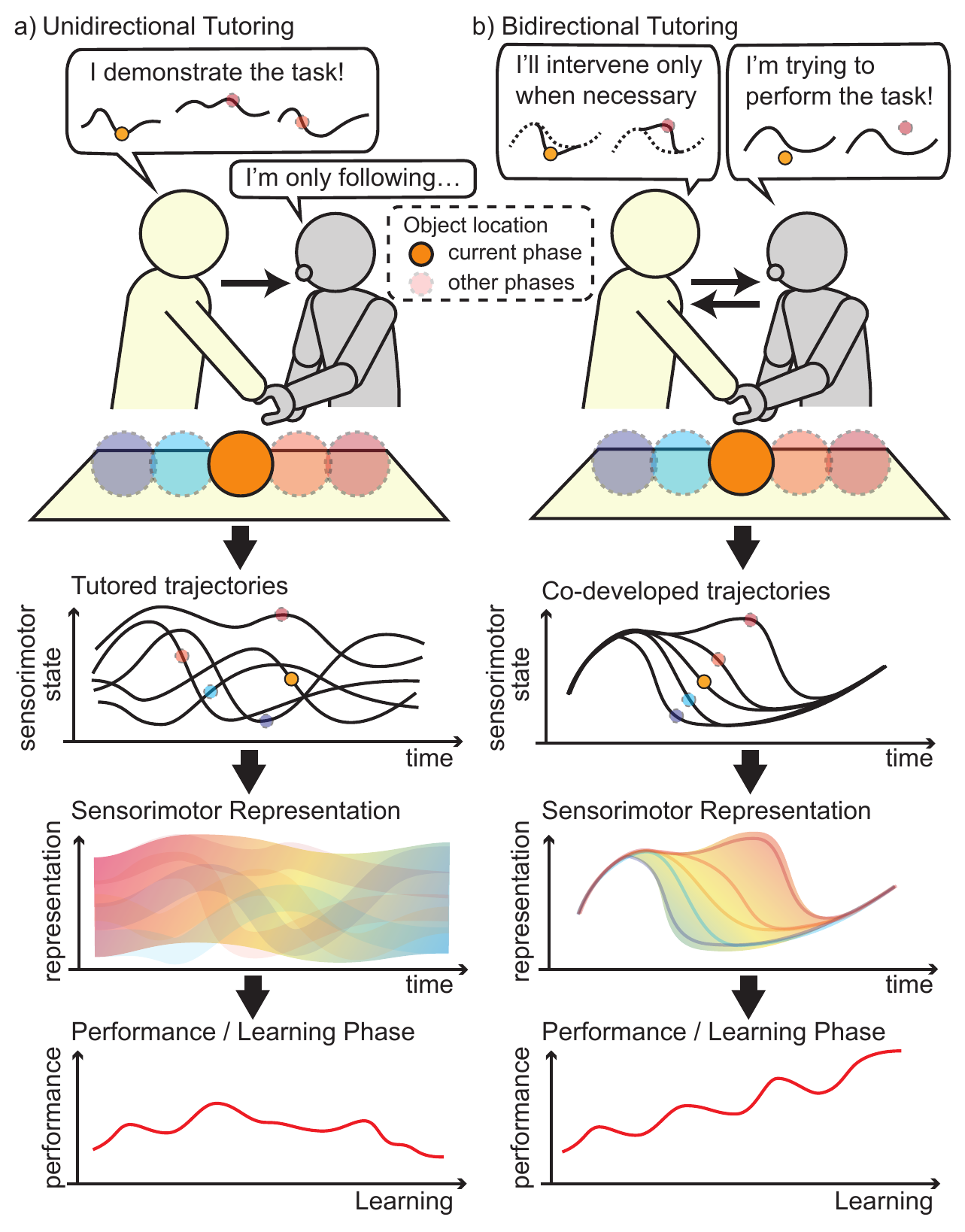}

    \caption{\textbf{Conceptual diagram of unidirectional and bidirectional tutoring in developmental motor learning.}
    The upper interaction schematics illustrate how tutoring trajectories are generated in each condition, and the middle panels overlay the resulting trajectories across developmental phases. The speech bubbles contrast unidirectional tutoring, where the tutor demonstrates the task while the robot passively follows, with bidirectional tutoring, where the robot attempts the task and the tutor intervenes as needed. Colored circles in the workspace indicate object locations across developmental phases; orange denotes the current phase, and the other colors denote other phases. In the trajectory panels, colored dots indicate task-critical states required for successful task execution, such as reaching the object location. The same colors are used in the trajectory and representation panels to indicate the corresponding object locations. The lower panels illustrate how these trajectories may be organized in the model's sensorimotor representation space. 
    \textbf{a} In unidirectional tutoring, tutor-provided demonstrations are illustrated as dispersed trajectories and as a poorly organized representation. 
    \textbf{b} In bidirectional tutoring, co-developed trajectories are illustrated as more aligned trajectories and as a more coherently organized representation.
    }    

    \label{fig:conceptual diagram} 
\end{figure}

Infants acquire motor and other cognitive skills through continuous and mutual interaction with caregivers~\cite{meltzoff1977imitation, pine2005tomasello, de2007participatory}. Such early social interactions are widely regarded as foundational to human developmental learning, where skills gradually emerge through guided embodied experiences~\cite{wood1976role, meltzoff2002imitative, lindblom2003social}. This socially grounded nature of development has been extensively discussed in developmental psychology, most notably Vygotsky’s theory~\cite{vygotsky1978mind, vygotsky2012thought}, which emphasizes that human development is scaffolded through social interaction. Inspired by this view, developmental robotics has also sought to incorporate social interaction into artificial learning systems~\cite{lungarella2003developmental, asada2009cognitive, cangelosi2018babies, ugur2015staged}. 
However, it remains unclear how mutual interaction between tutor and learner shapes the interaction dynamics themselves, and how these dynamics influence developmental motor learning in robots.

In robot motor learning, socially mediated skill acquisition is often formulated as learning from demonstration (LfD) or imitation learning (IL), where a robot acquires skills from tutor-provided examples~\cite{schaal1996learning, argall2009survey, schaal1999imitation}. 
A representative tutoring approach is kinesthetic teaching, in which a tutor physically guides the robot through task-relevant movements~\cite{haddadin2016physical, murata2013learning, wang2019learning}. 
Although kinesthetic teaching provides an intuitive way to demonstrate movements, it is often treated as a unidirectional process, in which the robot passively receives tutor-provided demonstrations~\cite{billard2006discriminative, kormushev2011imitation, ito2006dynamic}.
In such settings, new tutoring experiences are introduced without considering how the robot's ongoing behavior and developing internal dynamics may influence the demonstrated trajectories. 
Such a view may overlook how tutor--robot interaction dynamics shape tutoring experiences, including the structure of the resulting trajectories, and how these experiences in turn affect learning and generalization. Previous studies have shown that learning depends not only on whether individual demonstrations succeed, but also on the consistency of sensorimotor patterns across demonstrations. In particular, coherent sensorimotor patterns support stable behavioral representations, whereas poorly structured or inconsistent demonstrations can hinder learning and generalization~\cite{tani2003self, wang2025generalization, sakr2025consistency, shafiullah2022behavior}. This concern may become particularly relevant in developmental learning, where new tutoring experiences may be shaped through interaction with the robot's prior learning rather than introduced independently of its developing sensorimotor dynamics.

Building on this view, we hypothesize that bidirectional tutor--robot interaction supports stable developmental motor learning by preserving behavioral coherence across developmental phases. Fig.~\ref{fig:conceptual diagram} provides a conceptual overview of this hypothesis. 
In unidirectional tutoring (Fig.~\ref{fig:conceptual diagram}-a), tutor demonstrations are provided independently of the robot's behavior. Consequently, even when tutoring demonstrations satisfy task-critical states shown in Fig.~\ref{fig:conceptual diagram}, such as reaching the object and placing it at the target location, their detailed trajectories can still differ in shape and timing due to motor redundancy and natural variability across executions~\cite{latash2012bliss, van2004role, wu2014temporal}.
In this way, unidirectional tutoring may introduce successful but poorly structured demonstrations across developmental phases, which can hinder the formation of coherent sensorimotor representations and limit generalization.
In contrast, in bidirectional tutoring (Fig.~\ref{fig:conceptual diagram}-b), tutor intervention is closely coupled with the robot's ongoing behavior. Because this behavior reflects the robot's previously acquired sensorimotor dynamics, new trajectories are shaped jointly by tutor guidance and the robot's prior learning. As a result, co-developed trajectories can remain compatible with previously learned patterns while still allowing tutor intervention in new situations.

To examine this hypothesis, we conducted developmental learning experiments with a physical humanoid robot under two tutoring modes: \textit{unidirectional}, in which the robot passively receives guidance (Fig.~\ref{fig:conceptual diagram}-a), and \textit{bidirectional}, in which the tutor and robot co-develop behavior through real-time interaction (Fig.~\ref{fig:conceptual diagram}-b). We first performed a proof-of-concept experiment with a human tutor, conducted by one of the experimenters, who physically interacted with the robot. We then introduced an AI tutor to assess whether the observed effects of bidirectional tutoring would also emerge under a more controlled tutoring policy setting. The framework is implemented using a neural model based on the free energy principle (FEP)~\cite{friston2005theory, clark2015surfing, hohwy2013predictive}, specifically the Predictive coding-inspired Variational Recurrent Neural Network (PV-RNN)~\cite{ahmadi2019novel}. The same PV-RNN framework is also used to implement the AI tutor. As a framework for developmental learning, we extended PV-RNN with a generative replay mechanism to enable stable sequence-by-sequence learning without storing previous tutored sequences.

The main contributions of this study are summarized as follows:
\begin{itemize}
    \item We propose bidirectional tutoring as a developmental motor learning framework in which tutor intervention is coupled with the robot's ongoing behavior, allowing tutoring trajectories to be co-developed through tutor--robot interaction.

    \item We demonstrate, through human- and AI-tutoring experiments with a real robot, that bidirectional tutoring yields more coherent tutored trajectories and supports more stable phase-wise learning than unidirectional tutoring.

    \item We extend PV-RNN with generative replay to support stage-wise learning from single tutored sequences while mitigating catastrophic forgetting.

\end{itemize}

\section{Background}
\subsection{From Demonstration to Interactive Motor Learning}
Conventionally, kinesthetic teaching has been treated as a one-way process in which a robot passively receives tutor-provided demonstrations; for example, the robot’s motors may be inactivated during tutoring~\cite{tani2003self, billard2006discriminative, ito2006dynamic, kormushev2011imitation, borojevic2026deep}. Such settings correspond to unidirectional interaction, where adaptation is not required from either the tutor or the robot. In contrast, several studies have explored more interactive forms of kinesthetic teaching, allowing the robot to generate motion under compliant control during tutoring~\cite{tani2008codevelopmental, akgun2012trajectories, ikemoto2012physical, borojevic2026deep}. More recently, Matsumoto \etal~\cite{matsumoto2023incremental} advanced this direction by enabling online adaptation of robot motion in response to human physical guidance within an FEP-based active inference framework. In this framework, robot motion and tutor guidance were coupled during the acquisition of newly tutored episodes for incremental learning, marking a shift toward bidirectional tutoring. However, these studies did not directly examine how the bidirectional tutor-robot interaction shapes the interaction dynamics themselves, and how these dynamics, in turn, influence developmental motor learning. In contrast, the present study directly compares unidirectional and bidirectional tutoring to examine how the form of interaction shapes tutored trajectories and affects phase-wise developmental motor learning.

\subsection{Free Energy Principle}
The FEP provides a unifying framework for perception, action, and learning in biological systems~\cite{friston2005theory, clark2015surfing, hohwy2013predictive}. Under this principle, agents minimize a statistical quantity called variational free energy, which can be expressed as an upper bound of the surprise or negative logarithm of marginal likelihood, $-\ln{p(X)}$, where $X$ denotes the sensory observation. The variational free energy can be further decomposed into two components as follows:
\begin{align} 
    \underbrace{-\ln{p(X)}}_{\mathrm{surprise}} 
    &\leq \underbrace{\int{q(z|X)\ln{\frac{q(z|X)}{p(X, z)}}dz}}_{\mathrm{variational\ free\ energy}}\\
    &= \underbrace{D_{\mathrm{KL}}[q(z|X)||p(z)]}_{\mathrm{complexity}} - \underbrace{\mathbb{E}_{q(z|X)}[\ln{p(X|z)}]}_{\mathrm{accuracy}} \label{eq:variational free energy}
\end{align}
The first term represents complexity, penalizing deviation of the approximated posterior distribution $q(z|X)$ of the latent variable $z$ from the prior distribution $p(z)$. The second term represents accuracy, rewarding agreement between sensory prediction and observation.


The essential benefits for deployment of the FEP are twofold. First, its probabilistic formulation naturally handles the noise and stochasticity that are ubiquitous in embodied intelligent systems, as well as the subtle fluctuations arising from intentional conflicts during bidirectional interactions. Second, and more importantly, the FEP provides a mechanism for online inference by minimizing the mismatch between top-down priors and bottom-up posteriors. This allows the robot to continuously adjust its behavior in response to tutor intervention during real-time interaction~\cite{matsumoto2023incremental, sawada2024human}.

\begin{figure*}[tbph]
    \centering
    \includegraphics[width=0.99 \textwidth]{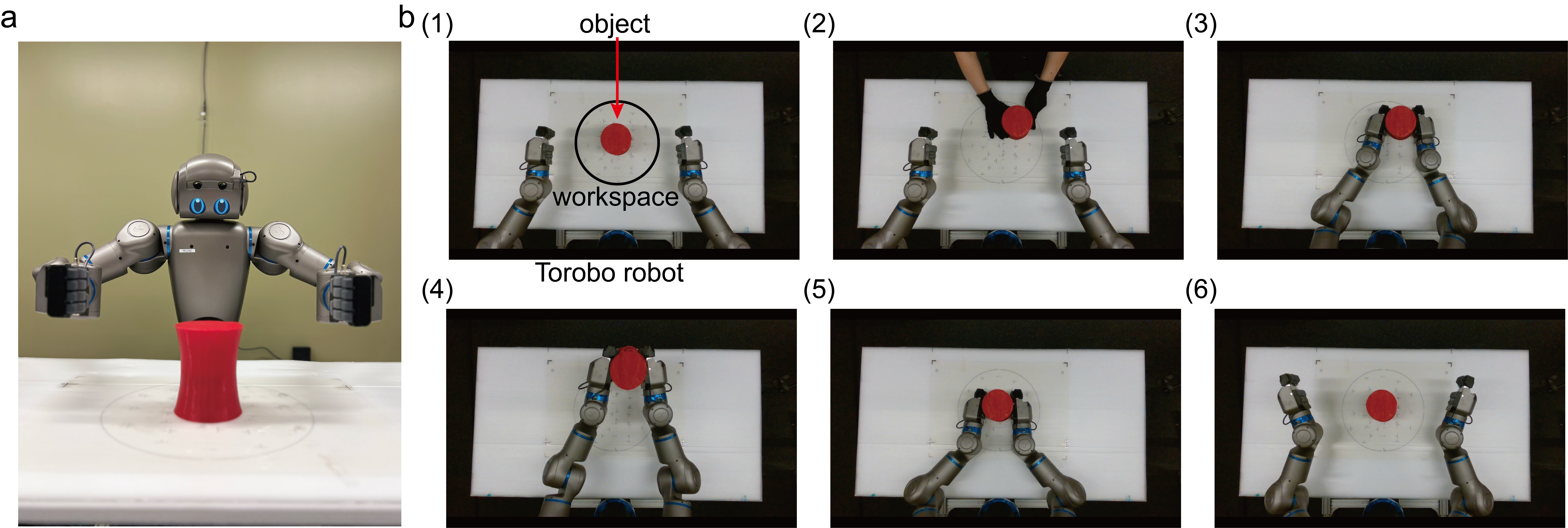}
    \caption{\textbf{Overview of experimental setups and task settings} \textbf{a} Torobo humanoid robot. \textbf{b} Example of a successful task sequence: 1) A robot begins in a home posture with a red cylindrical object placed at the center of the workspace indicated by a black circle; 2) the experimenter moves the red object to a random position; 3) the robot begins to reach for and grasp the object; 4) the robot lifts up the object; 5) the robot places and releases the object at the original location 6) the robot returns toward its initial posture.
    }
    \label{fig:experimental_setup}
\end{figure*}

\section{Methods} \label{chap:method}
In this section, we present our approach to investigating developmental motor learning under unidirectional and bidirectional interaction. We first describe the experimental setup and experimental procedure, and then introduce the computational framework underlying the two tutoring modes.

\subsection{Experimental Setup}
\subsubsection*{\textbf{Task Setting}}
To examine how different tutoring modes affect developmental motor learning, we designed a simple object manipulation task with a humanoid robot, Torobo (Tokyo Robotics) (Fig.~\ref{fig:experimental_setup}-a). The objective of the task is to reach and grasp a cylindrical red object approximately 12 cm in diameter placed at an arbitrary position within a 36 cm in diameter workspace, and then return it to the original position (Fig.~\ref{fig:experimental_setup}-b). 

\subsubsection*{\textbf{Torobo Humanoid Robot}}
The robot is equipped with a custom compliance module that allows humans to back-drive its joints during physical interaction. It has 6 joints for each arm and 2 joints for the torso. In the present experiment, the robot was controlled in Cartesian task space using a custom inverse kinematics module. The robot's proprioceptive sensations corresponded to the Cartesian poses of both end effectors, consisting of their positions and orientations. This proprioceptive representation contained 14 values in total: for each arm, 3 positional components $(x, y, z)$ and 4 quaternion components $(q_1, q_2, q_3, q_4)$ representing orientation. The visual sensation was represented by the two-dimensional center position of the object, extracted from overhead-camera images using an object-tracking program. Full pixel images were not used to simplify model learning and visual processing. All sensory and motor signals were sampled and processed at 10~Hz. Accordingly, each task episode consisted of a sequence of length $T{=}650$, corresponding to 65 seconds.

\subsection{Experimental Procedure and Evaluation Metrics}
\begin{figure}[t]
    \centering
    \includegraphics[width=0.48\textwidth]{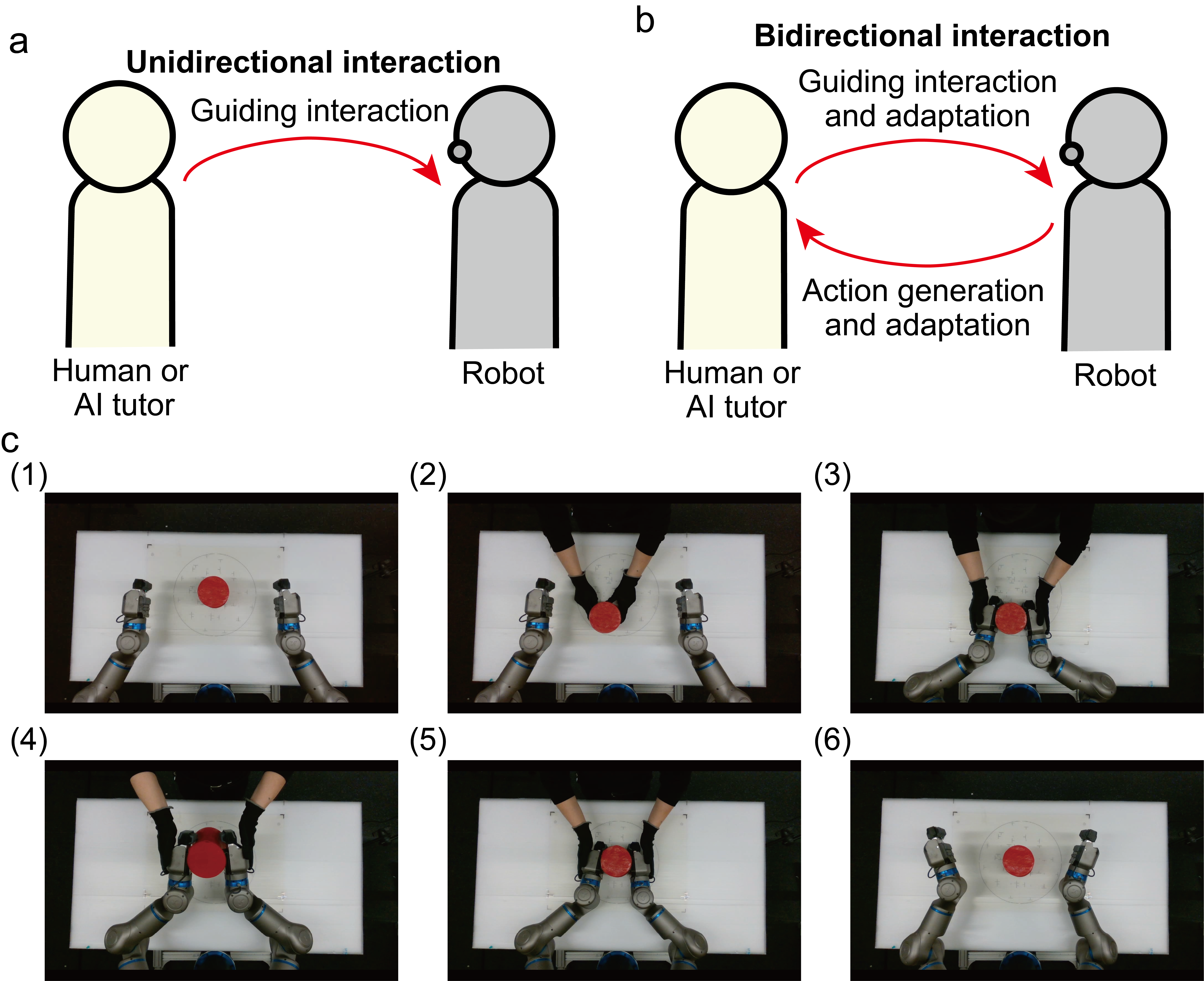}
    \caption{\textbf{Overview of tutoring setups} \textbf{a} Diagram of unidirectional tutoring, where the human or AI tutor provides one-way guidance to the robot. \textbf{b} Diagram of bidirectional tutoring, where the robot generates actions based on its own predictions while the tutor intervenes as needed. \textbf{c} Example of a human-robot tutoring sequence. 1-2) The object is moved to a random position; 3-5) the tutor provides physical guidance for the robot to reach for, grasp, lift, and return the object to the center; 6) the robot returns toward its initial posture.
    }
    \label{fig:tutoring_setup}
\end{figure}

In this study, we compared two tutoring modes: unidirectional tutoring (Fig.~\ref{fig:tutoring_setup}-a) and bidirectional tutoring (Fig.~\ref{fig:tutoring_setup}-b). In unidirectional tutoring, robot action generation was disabled, and the robot was set to gravity compensation mode so that the tutor could physically guide it through a successful task sequence. In bidirectional tutoring, the robot generated actions based on its own predictions under compliance mode, and the tutor intervened to correct the robot's behavior as necessary (Fig.~\ref{fig:tutoring_setup}-c). 

\subsubsection*{\textbf{Tutoring–Training Phase}} \label{section:tutoring_training}
The experiments progressed in a stage-wise manner through repeated tutoring–training phases. Each phase consisted of a single tutoring episode, followed by model training using the resulting tutored sequence. This process was repeated across ten phases. Phase~1 served as a common baseline for both conditions. Because the robot started from scratch and had not yet acquired any task-related behavior, bidirectional interaction was not feasible at this stage. Therefore, tutoring in Phase~1 was performed only in the unidirectional mode, and the resulting trajectory was used to initialize both learning conditions.

To ensure a statistically fair evaluation, we considered three sources of variability: the object positions and their learning-order across phases, human tutoring, and network stochasticity. To account for the first two, we repeated the same experimental procedure across three independently sampled object-position sets (Sets A--C). Each set consisted of 10 positions corresponding to the 10 developmental phases. To account for network stochasticity, we adopted the following multiple-seed training and selection procedure: (1) In Phase~1, the model was trained on the initial tutored trajectory using 15 different random seeds, accounting for both parameter initialization and stochastic sampling. (2) The trained models were sorted by their final loss, and the median-loss model was selected to avoid bias toward extreme outcomes. (3) The selected model was further trained with the newly obtained tutored sequence in 15 separate runs, each using a different random seed. (4) The trained models were again sorted by the final loss, and the median-loss model was selected for the next phase. Steps 3 and 4 were repeated through Phase~10. This procedure was applied independently to the experiment conducted with each object-position set (Sets A--C).

\begin{figure*}[tbph]
    \centering
    \includegraphics[width=0.99\textwidth]{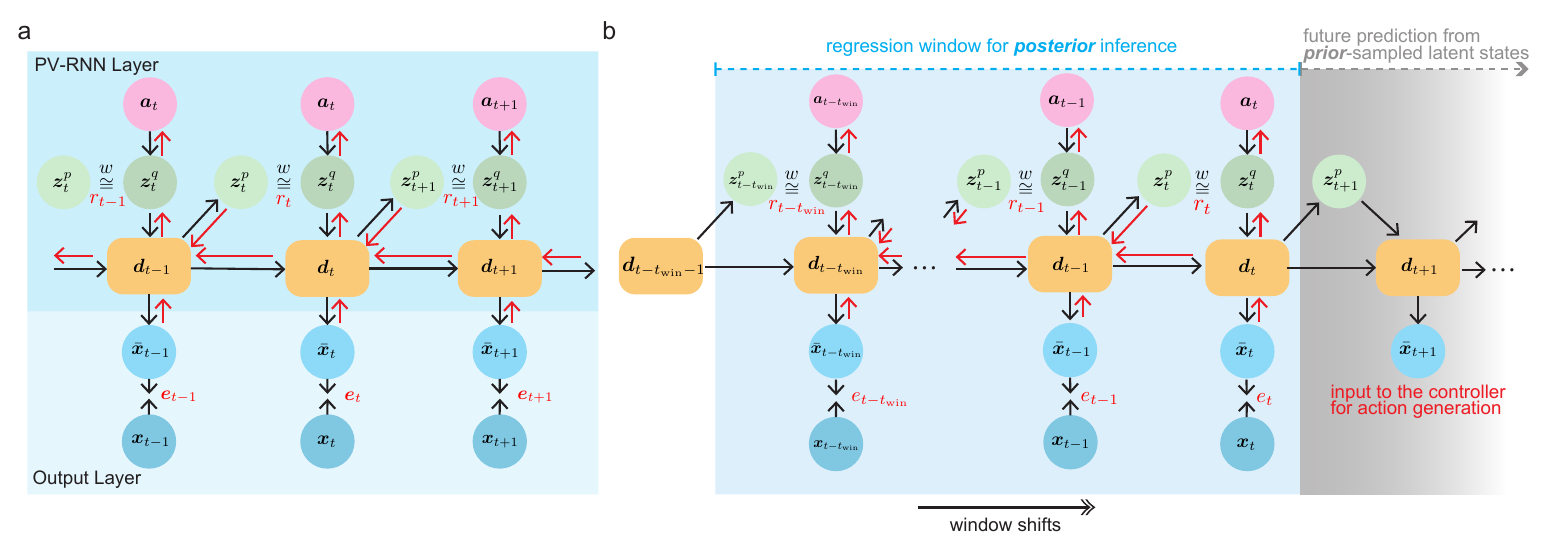}
    \vspace{-3mm}
    \caption{\textbf{Illustration of the computational flow of the PV-RNN during training and online inference.} \textbf{a} Training process. The model generates predictions over time through recurrent computation. At each time step, the reconstruction error $e_t$ is computed from the prediction $\bar{\bm{x}}_t$ and observation $\bm{x}_t$, while $r_t$  denotes the Kullback–Leibler divergence between the posterior $\bm{z}^{q}_{t}$ and prior  $\bm{z}^{p}_{t}$ latent distributions. \textbf{b} Online inference with a shifting regression window. Posterior estimates are updated within the window through error regression, and the resulting one-step-ahead prediction $\bar{\bm{x}}_{t+1}$ is used for action generation. As time advances, the regression window shifts forward, and the same procedure is repeated.
    Black arrows indicate forward computation, whereas red arrows indicate error backpropagation.
    }
    \label{fig:model}
\end{figure*}

\subsubsection*{\textbf{Testing Phase and Evaluation Metrics}} In the evaluation phase, the robot was expected to autonomously generate its own behavior without tutor assistance (Fig.~\ref{fig:experimental_setup}-b). To evaluate generalization performance, a separate test set consisting of 10 randomly sampled object positions was prepared, and each model was tested on this set at every stage ($N_{\mathrm{test}}=10$).

We evaluated model performance using the success rate, defined as $\frac{1}{N_{\mathrm{test}}}\sum_{i=1}^{N_{\mathrm{test}}}\mathbb{I}^{i}$, where $\mathbb{I}^i$ denotes the success indicator for the $i$-th test episode. The success indicator is defined as follows:
\begin{equation}
    \label{eq:reward_func}
    \mathbb{I}^{i} =
    \left\{
    \begin{array}{ll}
        1 &
        \begin{array}[c]{@{}l@{}}
            \text{if the episode was successful,} \\
            d_{\mathrm{reach}}^{i} < T_{\mathrm{err}}
            \text{ and }
            d_{\mathrm{place}}^{i} < T_{\mathrm{err}},
        \end{array}
        \\[12pt]
        0 & \text{otherwise.}
    \end{array}
    \right.
\end{equation}
Here, an episode was regarded as completed successfully if the robot finished the full manipulation sequence and released the object within the given episode length. Episodes in which the robot dropped the object or failed to complete the sequence within the episode length were counted as failures. The spatial error threshold $T_{\rm{err}}$ was introduced to evaluate the precision of the resulting behavior, so that success required not only task completion but also sufficient precision in reaching the object and placing it at the designated return location. This was necessary because the robot occasionally pushed the object with one hand before grasping it with both hands, or returned it to a position displaced from the designated return location. Therefore, we computed two spatial errors for each episode: the reaching error $d_{\mathrm{reach}}^{i}$ and the placement error $d_{\mathrm{place}}^{i}$. An episode was counted as successful only when both errors were smaller than the predefined threshold $T_{\rm{err}}$.

\subsection{Computational Framework}
In this section, we describe the computational framework used in this study, namely the Predictive coding–inspired Variational Recurrent Neural Network (PV-RNN). PV-RNN is grounded in the FEP, allowing the robot to continuously infer posterior latent variables by minimizing variational free energy, thereby supporting real-time behavioral adjustment. In the following section, we first introduce the model overview and training configuration, and then describe the online inference procedure used for real-time posterior inference and action generation, and finally present the generative replay method used in this study to support stage-wise developmental learning.

\subsubsection*{\textbf{Model Overview and Training Configuration}}
As illustrated in Fig.~\ref{fig:model}-a, the network consists of two main components: the PV-RNN layer and the output layer. At each time step $t$, the prediction $\bar{\bm{x}}_t$ is generated from the deterministic hidden state $\bm{d}_t$ and the stochastic latent variable $\bm{z}_t$. $\bm{a}_t$ is an adaptive vector used to infer the approximate posterior in each time step~\cite{ahmadi2019novel}. 
Here, these adaptive vectors are assigned in a sequence-specific manner, \ie each training sequence is associated with its own set of adaptive vectors. The output layer is implemented as a fully connected layer, which transforms the deterministic latent state $\bm{d}_t$ into sensorimotor prediction $\bar{\bm{x}}_t$. (See Appendix for the detailed computations.)

The objective function of the model is to minimize the variational free energy, equivalently maximizing the evidence lower bound (ELBO) $\mathcal{L}(\bm{\theta}, \bm{\phi})$ defined as follows:
\begin{equation}
    \begin{aligned}
        \mathcal{L}(\bm{\theta}, \bm{\phi}) = \sum_{t=1}^{T} \Biggl\{&\underbrace{\mathbb{E}_{q_{\bm{\phi}}(\bm{z}_{t})}\left[\ln p_{\bm{\theta}}(\bm{x}_{t} \mid \bm{d}_{t})\right]}_{\mathrm{Accuracy}:~e_t}\\
        &\quad - w\,\underbrace{D_{\mathrm{KL}}\left[q_{\bm{\phi}}(\bm{z}_{t})\;\|\;p_{\bm{\theta}}(\bm{z}_{t} \mid \bm{d}_{t-1})\right]}_{\mathrm{Complexity}:~r_t}\Biggr\}
    \label{eq:pv-rnn_variational_free_energy}
    \end{aligned}
\end{equation}
Here, $\bm{\theta}$ denotes the parameters of the generative model, which predicts sensory observations $\bm{x}_t$, $\bm{\phi}$ denotes the parameters of the inference model, which approximates the posterior distribution of latent variables given sensory observations. In the present PV-RNN model, $\bm{\theta}$ includes network weights, biases, and trainable initial hidden states $\bm{h}_0$, while $\bm{\phi}$ is realized as a set of adaptive vectors. 
$w$ is a hyperparameter called meta-prior, which balances the accuracy term and complexity term in the variational free energy~\cite{ahmadi2019novel}. 


The network was trained through backpropagation through time (BPTT)~\cite{amari2006theory}, and the optimization proceeds by gradient descent using the Adam optimizer with parameters $\alpha{=}0.01$, $\beta_{1}{=}0.9$, $\beta_{2}{=}0.999$~\cite{kingma2014adam}. The network parameters were set as follows: $N_{d}{=}60$, $N_{z}{=}1$, $\tau{=}8$, and $w{=}0.01$. Here, $N_{d}$ and $N_{z}$ denote the number of deterministic $\bm{d}$ and probabilistic $\bm{z}$ unit, respectively. The parameter $\tau$ defines the time constant for hidden-state computation, and $w$ corresponds to the meta-prior. The number of training epochs $N_{\mathrm{epoch}}$ in each phase was set to $30{,}000$.


\subsubsection*{\textbf{Online Inference and Action Generation}}
During both the tutoring and testing phases, PV-RNN performed online inference through an error regression (ER) process with a shifting window~\cite{ahmadi2019novel, ohata2025characterizing}. The overall computational flow is illustrated in Fig.~\ref{fig:model}-b (see Appendix for detailed computations). In this process, the network dynamically adjusted its inference parameters $\bm{\phi}$ based on the mismatch between sensory observations and predictions, while keeping the generative parameters $\bm{\theta}$ fixed. Specifically, $\bm{\phi}$ was repeatedly optimized by minimizing the following variational free energy:
\begin{equation}
    \begin{split}
        \mathcal{L}(\bm{\phi}) = \sum_{t'=t-t_{\mathrm{win}}}^{t} \Biggl\{   &\underbrace{\mathbb{E}_{q_{\bm{\phi}}(\bm{z}_{t-t_{\mathrm{win}}:t'})}\left[\ln p_{\bm{\theta}}(\bm{x}_{t'} \mid \bm{d}_{t'})\right]}_{\mathrm{Accuracy}:~e_{t'}}\\
        &\quad - w\,\underbrace{D_{\mathrm{KL}}\left[q_{\bm{\phi}}(\bm{z}_{t'})\;\|\;p_{\bm{\theta}}(\bm{z}_{t'} \mid \bm{d}_{t'-1})\right]}_{\mathrm{Complexity}:~r_{t'}}\Biggr\}
    \label{eq:pv-rnn_variational_free_energy}
    \end{split}
\end{equation}
ER was performed using BPTT within a regression window of length $t_{\mathrm{win}}$ for $N_{\mathrm{itr}}$ optimization steps. In this study, we set $t_{\mathrm{win}}{=}100$ and $N_{\mathrm{itr}}{=}50$. The same Adam optimizer settings as in training were used. After optimization, the one-step-ahead prediction $\bar{\bm{x}}_{t+1}$ was generated, and its proprioceptive components were sent to the robot controller for action generation. The regression window was then shifted forward as time advanced, and the same procedure was repeated.

\subsubsection*{\textbf{PV-RNN-Based Generative Replay for Stage-Wise Learning}}

\begin{algorithm}[t]
    \caption{Generative replay with PV-RNN}
    \label{alg:generative_replay}
    \begin{algorithmic}[1]
        \Require New tutored sequence $\bm{X}_{\mathrm{new}}=\{\bm{x}_{t}\}^{T}_{t=1}$ and its inferred adaptive vectors $\bm{A}_{\mathrm{new}}=\{\bm{a}_{t}\}^{T}_{t=1}$
        \State Generate replay sequences from prior-sampled latent states of the current model, and obtain the corresponding prior parameters:
            \[
            \bar{\bm{X}}_j=\{\bar{\bm{x}}_{t,j}\}_{t=1}^{T},\ \{(\bm{\mu}^{p}_{t, j}, \bm{\sigma}^{p}_{t, j})\}^T_{t=1} \qquad j=1,\dots,N_{\mathrm{replay}}
            \]
        \State Initialize the corresponding adaptive vectors from the prior parameters:
            \[
            \bm{a}^{\mu}_{t,j}\leftarrow \mathrm{arctanh}(\bm{\mu}^{p}_{t,j}),\quad
            \bm{a}^{\sigma}_{t,j}\leftarrow \mathrm{logit}(\bm{\sigma}^{p}_{t,j})
            \]
            \[
            \bm{A}_j=\{(\bm{a}^{\mu}_{t,j},\bm{a}^{\sigma}_{t,j})\}^{T}_{t=1} \qquad j=1,\dots,N_{\mathrm{replay}}
            \]
        \State $\mathcal{M}\leftarrow \{(\bar{\bm{X}}_j,\bm{A}_j)\}_{j=1}^{N_{\mathrm{replay}}}$
        \For{$N_{\mathrm{epoch}}$}
            \State $\mathcal{B}\leftarrow \{(\bm{X}_{\mathrm{new}},\bm{A}_{\mathrm{new}})\}\cup\{(\bar{\bm{X}}_{k},\bm{A}_{k})\}_{k=1}^{N_{\mathcal{B}}-1}\sim\mathcal{M}$
            \State Update $\bm{\theta}$ and $\bm{\phi}$ using one epoch of gradient descent on $\mathcal{L}(\bm{\theta},\bm{\phi};\mathcal{B})$
        \EndFor
    \end{algorithmic}
\end{algorithm}

To support stage-wise learning, we developed a generative replay mechanism tailored to the PV-RNN learning framework~\cite{tani1998, french1999, shin2017continual}. 
In this process, replay sequences were generated by sampling latent states from the prior distribution of the trained generative model while recursively updating the recurrent dynamics forward in time. Because these sequences reflect previously acquired sensorimotor patterns, replaying them during later training helps mitigate catastrophic forgetting without storing the previous training data~\cite{mccloskey1989catastrophic, french1999catastrophic}.

In the present stage-wise learning setting, the model needed to incorporate a newly tutored sequence at each phase while preserving sensorimotor patterns acquired in previous phases. Because only one new tutored sequence was available at each phase, standard mini-batch training with replay samples could easily result in either overfitting or forgetting. Too many replay samples could bias training toward previously acquired sensorimotor patterns, whereas too few replay samples could provide insufficient information about those patterns and lead to forgetting. To address this issue, we generated a replay buffer containing many replay-sequences, but used randomly sampled small subset of them at each training epoch. This allowed the model to repeatedly revisit past sensorimotor patterns without overwhelming the newly tutored sequence. Algorithm~\ref{alg:generative_replay} provides the pseudo-code for this generative replay procedure used in this study. After each tutoring episode, the resulting tutored sequence was used as the new training sequence $\bm{X}_{\mathrm{new}}{=}\{\bm{x}_t\}^{T}_{t=1}$. Before training on this sequence, a replay buffer $\mathcal{M}$ was constructed from $N_{\mathrm{replay}}$ replay sequences generated from the current trained model. 
In addition, because PV-RNN requires sequence-specific adaptive vectors, adaptive vectors were prepared for both the new tutored sequence and replay sequences. Although adaptive vectors are randomly initialized in standard PV-RNN training~\cite{ahmadi2019novel}, random initialization was less stable in this incremental learning setting because it ignored sequence-specific latent information. We therefore initialized the adaptive vectors as follows. For the new tutored sequence, the adaptive vectors inferred during the tutoring session were used as $\bm{A}_{\mathrm{new}}$. For each replay sequence, the adaptive vectors $\bm{A}_j$ were initialized using the sequence-specific stochastic parameters $\{(\bm{\mu}^{p}_{t,j}, \bm{\sigma}^{p}_{t,j})\}_{t=1}^{T}$ obtained during its replay generation. This allowed both the new and replayed sequences to be initialized with adaptive vectors consistent with their corresponding latent dynamics before training. Here, $\mathrm{arctanh}(\cdot)$ and $\mathrm{logit(\cdot)}$ were used as inverse transformations of those used to compute the posterior parameters $\bm{\mu}^{q}$ and $\bm{\sigma}^{q}$, respectively (see Appendix). Finally, at each training epoch, the model is updated using the training batch $\mathcal{B}$ of size $N_\mathcal{B}$, which consisted of the new tutored sequence and a randomly sampled subset of replay sequences. In the present experiments, $N_{\mathrm{replay}}{=}1023$ and $N_{\mathcal{B}}{=}8$ were used.

\section{Human Tutor Experiment: Proof of Concept}
As an initial proof of concept, we first conducted a developmental motor learning experiment in which a human tutor physically interacted with a humanoid robot. This experiment examined whether the proposed bidirectional tutoring process can arise in real physical human--robot interaction, where both tutor and robot are embodied agents coupled through physical contact and real-time sensorimotor feedback. Testing multiple human tutors in this setting is practically difficult because each participant would need to complete the full multi-stage procedure, requiring substantial experimental time. In the present proof-of-concept experiment, one of the experimenters served as the human tutor. For this reason, the human-tutoring experiment is positioned as an embodied demonstration of the proposed mechanism, which was further evaluated under more controlled conditions using an AI tutor in Section~\ref{section:ai_tutor}.

\begin{figure}[t]
    \centering
    \includegraphics[width=0.5\textwidth]{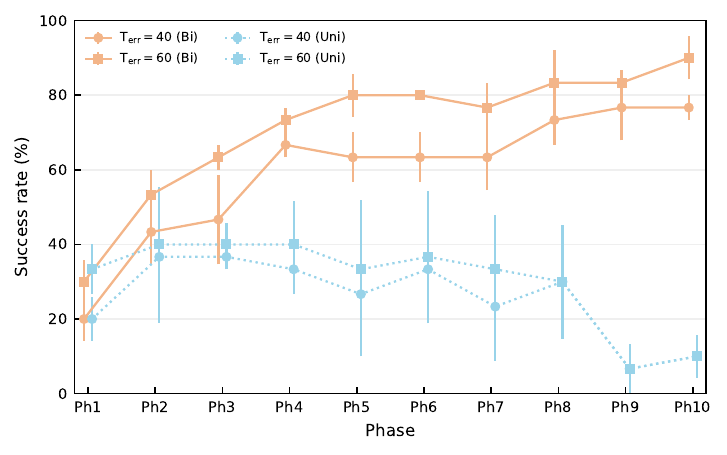}
    \vspace{-8mm}
    \caption{\textbf{Success rate across developmental phases for the unidirectional (Uni) and bidirectional (Bi) tutoring conditions.} Data points represent the mean success rate across training sets A--C, and error bars indicate the standard error of the mean (SEM). Two spatial error thresholds are used, $T_{\mathrm{err}}{=}40$ and $60$.
    }
    \label{fig:human_sr}
\end{figure}

\subsection{Task Performance}
First, we evaluated task performance across ten training phases by computing the success rate under two spatial error thresholds ($T_{\mathrm{err}}{=}40$ and $60$ pixels; $1$~pixel${\approx}1.088$~mm), where the smaller threshold indicates the stricter success criterion. Fig.~\ref{fig:human_sr} shows the phase-wise task performance for the unidirectional and bidirectional tutoring conditions. Overall, the bidirectional condition consistently achieved higher success rates than the unidirectional condition across phases under both error thresholds. In the bidirectional condition, the success rates increased progressively over developmental phases, indicating gradual improvement in task performance through stage-wise learning. In contrast, the unidirectional condition remained substantially lower overall and did not exhibit a clear improvement trend. Instead, performance plateaued in the early phases and declined in the later phases. The gap between the two conditions became particularly pronounced in the later phases, where the mean success rate in the bidirectional condition reached $90.0\%$ under $T_{\mathrm{err}}{=}60$ and $76.7\%$ under $T_{\mathrm{err}}{=}40$, whereas that in the unidirectional condition remained at $10.0\%$ under both thresholds. These results indicate that bidirectional tutoring better supported stage-wise motor learning, whereas unidirectional tutoring showed limited improvement and a decline in later phases. Representative autonomous testing episodes are shown in \href{https://youtu.be/cgVFxeJdcck}{Supplementary Video~1}, illustrating the difference in task execution between the unidirectional and bidirectional conditions.

\begin{figure*}[tbph]
    \centering
    \includegraphics[width=0.99\textwidth]{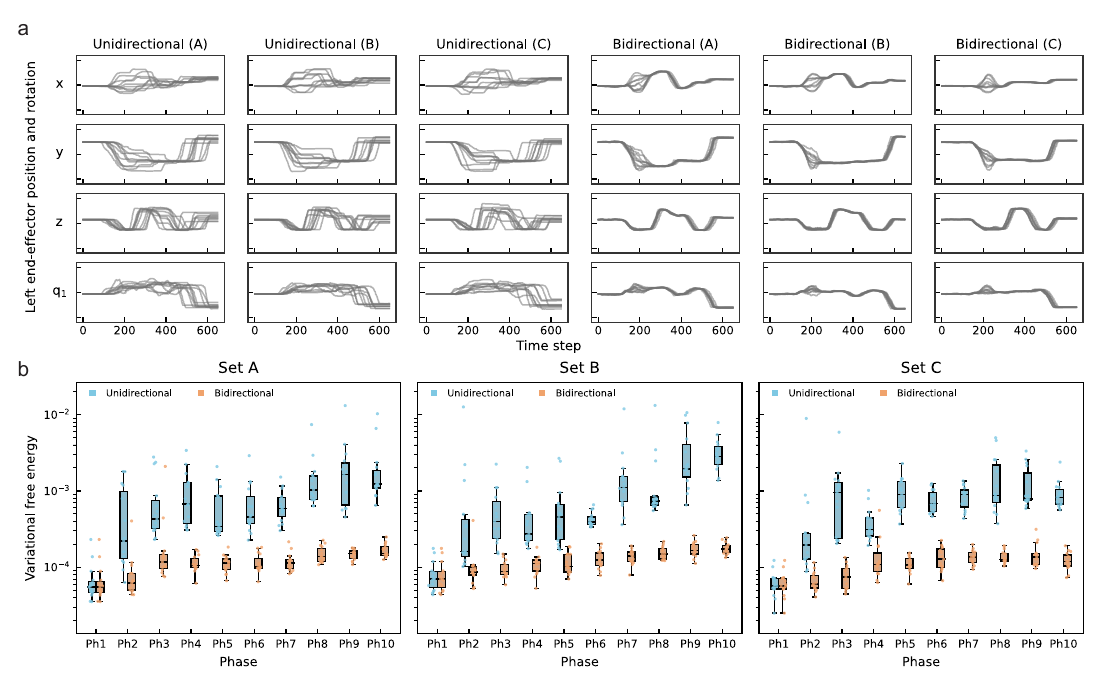}
    \vspace{-3mm}
    \caption{\textbf{Qualitative and quantitative comparison of tutoring dynamics under the unidirectional and bidirectional conditions.} 
    \textbf{a} Left end-effector position and rotation trajectories during tutoring, overlaid across all tutoring phases for each training set (A--C). Each trace corresponds to the tutored sequence at one phase. For rotation, only the first quaternion component ($q_1$) is shown, with the other components ($q_2$, $q_3$, and $q_4$) omitted for clarity. 
    \textbf{b} Training loss, quantified as variational free energy, across developmental phases for each training set (A--C). Box plots show the distribution across random seeds. Statistical analysis was conducted on log-transformed values from Phases 2--10. Phase 1 was excluded from the statistical analysis because it served as a common baseline before the tutoring conditions diverged.
    }
    \label{fig:human_traj_loss}
    \vspace{-3mm}
\end{figure*}

\subsection{Qualitative and Quantitative Comparison} 
To better understand the factors underlying the difference in task performance, we examined the qualitative and quantitative characteristics of the tutored trajectories. In the present stage-wise learning procedure, each developmental phase introduced a new tutored trajectory that was used for subsequent model training. Therefore, the structure of these trajectories was expected to affect how compatible new tutoring experiences were with the model's previously acquired sensorimotor patterns. We first examined the spatiotemporal organization of the tutored trajectories across developmental phases. We then analyzed the training loss to assess whether differences in trajectory organization were reflected in the model's learning dynamics. To provide an intuitive view of the tutoring process, representative tutoring episodes are shown in \href{https://youtu.be/GxxSs259urM}{Supplementary Video~2}. The video highlights differences in how tutored trajectories were formed under the two tutoring modes and shows the robot's predictions, observations, internal states, and tutor-applied excess torque.

\subsubsection*{\textbf{Spatiotemporal Structure of Tutored Trajectories}}
We first visualized the trajectories obtained during tutoring sessions across the ten developmental phases. Fig.~\ref{fig:human_traj_loss}-a overlays the left end-effector position and rotation trajectories obtained during tutoring across all developmental phases under unidirectional and bidirectional conditions. In the unidirectional condition, the trajectories exhibited substantial spatial and temporal variability across all dimensions. In contrast, in the bidirectional condition, the trajectories showed stronger spatiotemporal consistency across developmental phases. Importantly, this consistency was expressed differently depending on the position and rotation components. The $x$- and $y$-position trajectories showed spatial dispersion around $t{=}200$, corresponding to the reaching movement toward different object locations. This dispersion likely reflects task-critical adjustments, where the robot had to approach newly introduced object locations under tutor guidance. Outside these task-critical adjustments, the $x$- and $y$-position trajectories became more aligned in both shape and timing across phases. The vertical position trajectory ($z$) and rotation trajectory (shown here by the representative quaternion component $q_1$) also showed clear temporal consistency across phases. This consistency likely reflects the fact that these movement components required less adjustment and could therefore rely more on sensorimotor patterns acquired in previous phases. Together, these observations suggest that bidirectional tutoring did not simply reduce variability uniformly. Rather, the resulting trajectories reflected both tutor-guided adjustments to newly introduced object locations and movement patterns previously acquired by the robot.

\subsubsection*{\textbf{Training Loss and Trajectory Compatibility}}
We next examined the training loss, quantified as the variational free energy, across developmental phases. In the present context, the loss value can be interpreted as an index of how compatible the tutored trajectories were with the model's previously acquired sensorimotor dynamics. 
Higher loss suggests that the trajectories were harder to predict and integrate with prior sensorimotor patterns, likely because they introduced greater variability or spatiotemporal inconsistency. Lower loss, in contrast, may indicate that the trajectories were easier to predict and integrate because they were more compatible with previously acquired sensorimotor dynamics.


As shown in Fig.~\ref{fig:human_traj_loss}-b, the bidirectional condition generally yielded lower training loss than the unidirectional condition across developmental phases and training sets. This difference remained significant in a linear model that accounted for developmental phase and training set (condition effect for bidirectional tutoring relative to unidirectional tutoring: $\beta{=}{-}1.818$, SE${=}0.044$, $t{=}{-}41.01$, $p{<}0.001$). Although loss increased overall across phases in both tutoring conditions, this trend mainly reflects the cumulative learning procedure, in which a newly tutored sequence was added at each phase and expanded the range and complexity of sensorimotor patterns that the model had to represent. Overall, the lower loss in the bidirectional condition suggests that its tutored trajectories were more compatible with the model's previously acquired sensorimotor dynamics than those obtained under unidirectional tutoring.

\begin{figure}[t]
    \centering
    \includegraphics[width=0.48\textwidth]{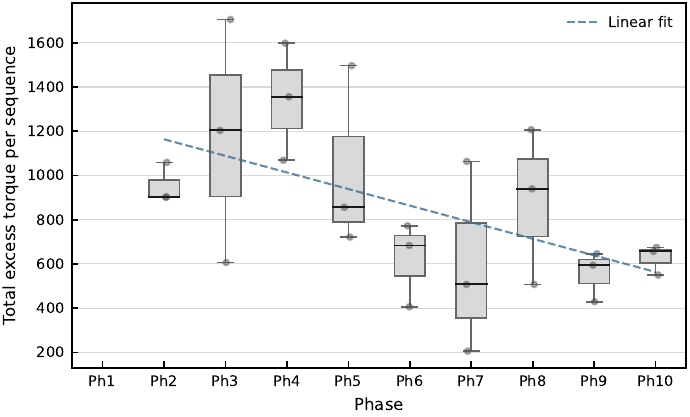}
    \vspace{-2mm}
    \caption{\textbf{Changes in human tutor intervention during bidirectional tutoring.} Box plots show the total excess torque applied by the human tutor during each tutoring episode in the bidirectional condition. Data from learning sets A–C were pooled within each developmental phase. The dashed line indicates the linear trend estimated by linear regression across developmental phases. Total excess torque decreased over phases (Pearson's correlation, $r{=}{-}0.726$, $p{=}0.0268$; linear regression, $R^{2}{=}0.527$). No value is shown for Ph1 because this baseline phase did not include robot-to-tutor physical feedback, which was required to compute this measure.
    }
    \label{fig:human_torque}
    \vspace{-3mm}
\end{figure}

\subsubsection*{\textbf{Reduction of Tutor Intervention across Developmental Phases}}
In addition to the qualitative and quantitative characteristics of the tutored trajectories, we examined the total amount of tutor intervention across developmental phases in the bidirectional condition as an additional behavioral index of learning dynamics. As shown in Fig.~\ref{fig:human_torque}, the total excess torque showed an overall decreasing trend across developmental phases. This reduction suggests that the robot gradually required less external guidance from the tutor across developmental phases. Together with the increased task success rate observed in Fig.~\ref{fig:human_sr}, this decrease in tutor intervention suggests that the robot acquired task-relevant motor skills while becoming less dependent on tutor guidance.

\section{AI tutor Experiment: Controlled Validation} \label{section:ai_tutor}
The human-tutoring experiment provided an initial proof of concept that bidirectional tutoring can support stable developmental learning through the formation of consistent behavioral patterns. To examine this mechanism under more controlled conditions, we implemented an AI tutor that captured selected functional aspects of human tutoring. Using the same AI tutor in both conditions allowed us to examine the effect of interaction mode while keeping the tutoring policy fixed. 

\subsection{Training the PV-RNN-Based AI Tutor}\label{sec:training_ai_tutor}
\begin{figure}[b]
    \centering
    \includegraphics[width=0.50\textwidth]{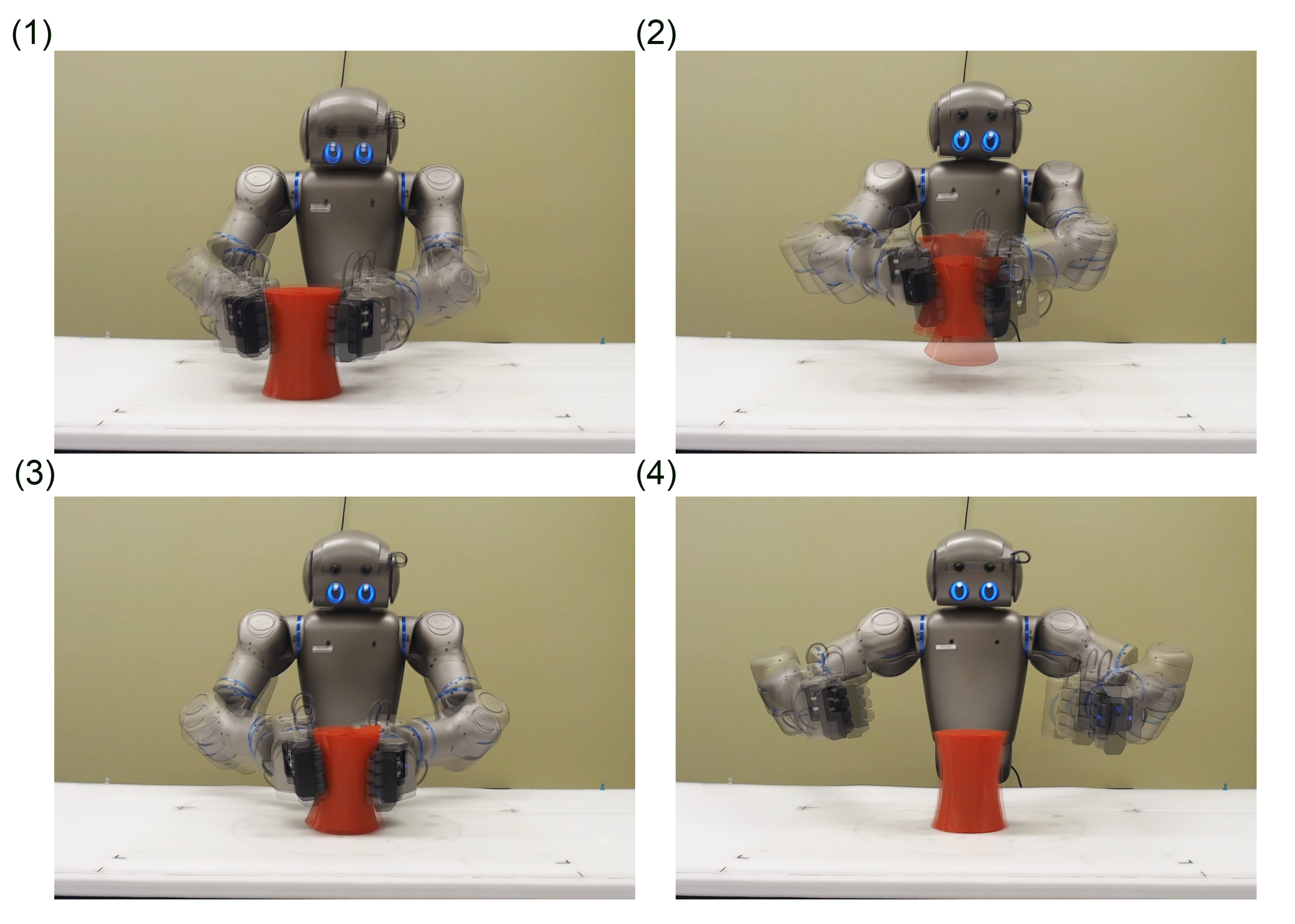}
    \caption{\textbf{Example of variability in AI-tutor-generated trajectories under the same object-location condition.} Panels (1)--(4) show overlaid snapshots from three repeated executions at representative time points: (1) reaching and grasping at $t{=}200$, (2) lift-up at $t{=}400$, (3) put-down at $t{=}500$, and (4) returns toward its initial posture at $t{=}650$. The snapshots illustrate that the trained AI tutor generated task-relevant movements with variability, rather than producing a single deterministic trajectory.
    }
    \label{fig:ai_tutor_example}
\end{figure}
We first describe how the AI tutor model was trained. The AI tutor was designed to provide a controlled tutoring policy for examining the effect of interaction mode while minimizing the direct influence of a particular human tutor. Therefore, instead of training the AI tutor on recorded human demonstrations, we generated the training trajectories computationally. Furthermore, these trajectories were designed to preserve the task-relevant structure while incorporating spatiotemporal variability. This avoided purely deterministic training examples and allowed the AI tutor to learn variability inherent in embodied motor behavior. To cover different object-location conditions and possible movement variations, we generated 555 trajectories and used them to train the AI tutor model. Details of the training-trajectory generation procedure are provided in Supplementary Material.

The AI tutor model was implemented using a PV-RNN model for two reasons. First, as a probabilistic recurrent model, PV-RNN is well suited for learning and generating stochastic time-series data~\cite{murata2013learning}, allowing the AI tutor to capture variability in the training trajectories. Second, the PV-RNN framework supports online inference, allowing the AI tutor to update its prediction in response to the robot's ongoing movement during bidirectional interaction. This online update was used to adjust tutor intervention, as described in the next section (Section~\ref{sec:intervention_weight}). The network parameters of the AI tutor model were set to $N_d{=}100$, $N_z{=}15$, $\tau{=}16$, $w{=}0.01$, and $N_{\mathrm{epoch}}{=}100{,}000$. Fig.~\ref{fig:ai_tutor_example} shows three repeated generations by the trained AI tutor under the same object-location condition, illustrating variable movement execution while preserving the task-relevant sequence structure. This trained AI model was used in the following AI-tutoring experiments.



\begin{figure*}[tbhp]
  \centering
  \includegraphics[width=0.98\textwidth]{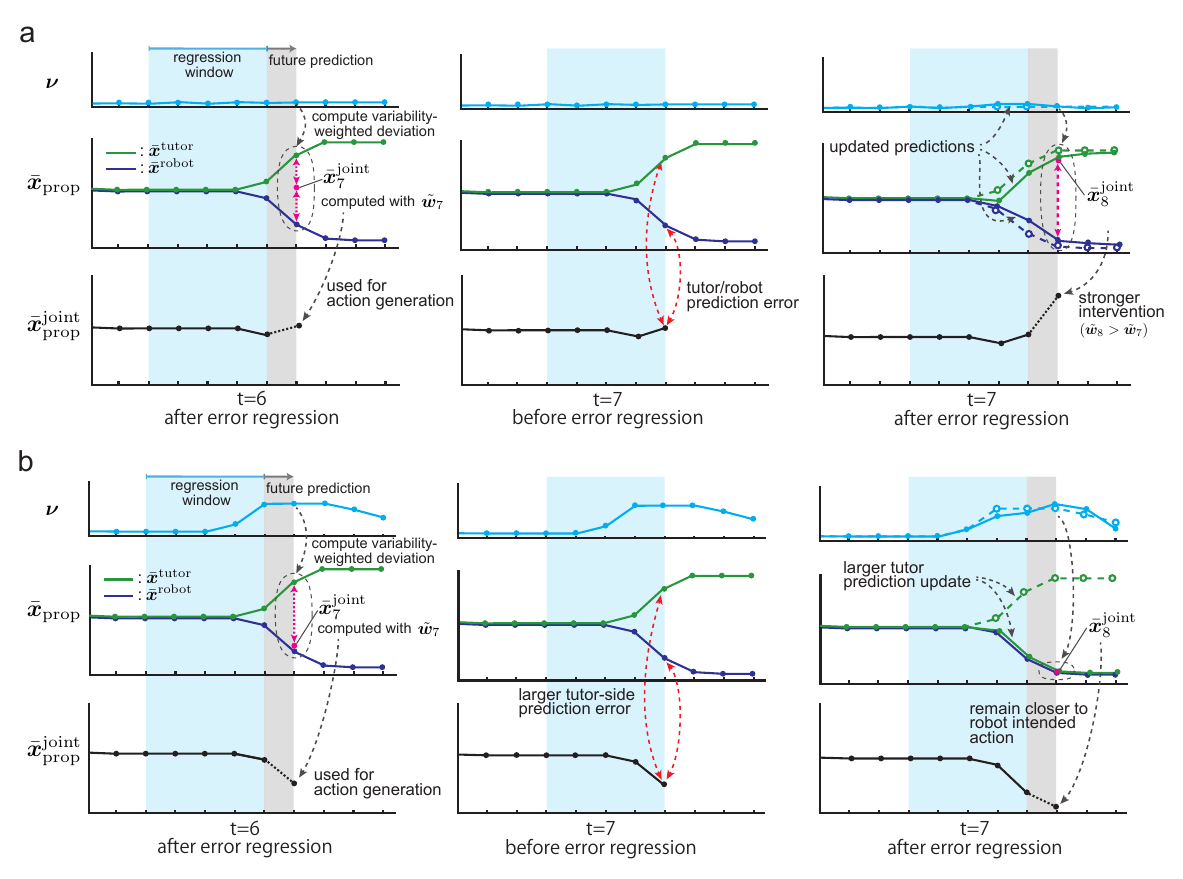}
  \vspace{-4mm}
  \caption{\textbf{Schematic illustration of AI-tutor intervention through online inference in the bidirectional condition.} (a) Low expected movement variability case, where $\nu$ is small in the AI tutor model. (b) High expected movement variability case, where $\nu$ is large in the AI tutor model. At $t{=}6$ (left), after observing $\bm{x}_{1:6}$, the tutor and the robot generate their predicted actions, and the joint action is computed from these predictions. For a similar deviation between the robot's intended action and the tutor's prediction, the low-variability case yields a larger intervention weight $\tilde{\bm{w}}$, so the joint action is drawn closer to the tutor's intended action. In contrast, when $\nu$ is larger, the intervention weight becomes smaller, and the joint action remains closer to the robot's intended action. At $t{=}7$ before error regression (middle), prediction error is computed from the mismatch between the predicted trajectory $\bar{\bm{x}}_{3:7}$ and the observation $\bm{x}_{3:7}$. Because weaker intervention in the high-variability case makes the generated observation farther from the tutor's prediction, the tutor-side prediction error becomes larger than in the low-variability case. At $t{=}7$ after error regression (right), both predictions are updated based on their respective prediction errors. In the low-variability case, the next joint action $\bar{\bm{x}}^{\mathrm{joint}}_{8}$ is pulled further toward the tutor's prediction because the updated intervention weight becomes larger ($\tilde{\bm{w}}_{8}>\tilde{\bm{w}}_{7}$). In the high-variability case, the tutor prediction is updated more strongly, resulting in the joint action remaining closer to the robot's intended action.  
  }
  \label{fig:ai_tutor}
  \vspace{-2mm}
\end{figure*}

\subsection{Bidirectional Interaction between AI tutor and Robot} \label{sec:intervention_weight}
We next describe how the AI tutor interacted with the real robot and adjusted its intervention based on the robot's ongoing behavior. The AI tutor was not intended to reproduce full human tutoring behavior. Instead, it was designed as a controlled intervention policy that allowed us to isolate whether coupling the tutor intervention to the robot’s ongoing behavior is sufficient to produce coherent developmental trajectories. 

In the unidirectional condition, the AI tutor generated a tutoring trajectory based on its learned dynamics, and the robot passively followed this trajectory without autonomous action generation. In the bidirectional condition, by contrast, the robot generated its own intended action from its learned dynamics, while the AI tutor intervened as needed. To implement such bidirectional interactions, the AI tutor needed to determine when and how strongly to intervene based on the robot's ongoing behavior. We formulated this intervention process using the tutor's expected movement variability, so that tutor intervention could be adjusted according to the context of the ongoing movement. The following subsections describe this mechanism in four steps: 1) estimation of the tutor's expected movement variability, 2) variability-weighted deviation computation, 3) intervention-weight computation, and 4) joint-action generation.

\subsubsection*{\textbf{1) Estimation of the Tutor's Expected Movement Variability}} 
First, we estimated the tutor's expected movement variability for the upcoming movement. This quantity serves as a context-dependent tolerance for interpreting the deviation between the robot's intended action and the AI tutor's predicted action. Because the predicted variance of the prior latent distribution $(\bm{\sigma}^{p,\mathrm{tutor}}_{t+1})^2$ represents uncertainty in the tutor's next-step prediction, we used it as a proxy for this expected variability. Specifically, we computed
\begin{equation}
   \nu_{t+1} = \frac{1}{N_z}\sum^{N_z}_{k=1}{(\bm{\sigma}^{p, \mathrm{tutor}}_{t+1, k})^2}
\end{equation}
where $\nu_{t+1}$ denotes the tutor's expected movement variability for the upcoming movement, computed as the mean predicted latent variance across the $N_z$ stochastic units.

\subsubsection*{\textbf{2) Variability-weighted Deviation Computation}} 
Second, we computed the deviation between the robot's intended action and the AI tutor's predicted action relative to the expected variability estimated in Step 1. This allowed the same absolute action difference to be interpreted in a context-dependent manner: a deviation was treated as more significant when the tutor expected a narrow range of possible movements, and less significant when the tutor expected a broader range of possible movements. For each proprioceptive dimension, the variability-weighted deviation was computed as
\begin{equation}
\bm{\delta}_{t+1, i}  = \frac{1}{\nu_{t+1}}\left| \bar{\bm{x}}^{\mathrm{robot}}_{t+1, i} - \bar{\bm{x}}^{\mathrm{tutor}}_{t+1, i}\right|
\end{equation}
where $\bar{x}^{\mathrm{robot}}_{t+1,i}$ and $\bar{x}^{\mathrm{tutor}}_{t+1,i}$ denote the $i$-th proprioceptive dimensions of the robot's intended action and the AI tutor's predicted action, respectively.


\subsubsection*{\textbf{3) Intervention Weight Computation}} 
Third, we converted the variability-weighted deviation into an intervention weight. This mapping was designed to satisfy three practical requirements: the AI tutor should not intervene for deviations within a noise threshold, the intervention strength should increase gradually as the deviation becomes larger, and the resulting weight should remain bounded between 0 and 1. We implemented this mapping using a simple bounded function that satisfied these requirements as follows:
\begin{equation}
\label{eq:intervention_weight}
\tilde{\bm{w}}_{t+1, i} =
\begin{cases}
1 - \exp(-\lambda \bm{\delta}_{t+1, i}) & \text{if } \bm{\delta}_{t+1, i} > \epsilon_{\mathrm{noise}},\\
0 & \text{otherwise}.
\end{cases}
\end{equation}
where $\tilde{w}_{t+1,i}$ denotes the intervention weight for the $i$-th proprioceptive dimension. The parameter $\lambda$ controls how rapidly the intervention weight increases, and $\epsilon_{\mathrm{noise}}$ defines a noise threshold. In this study, we set $\lambda{=}10$ and $\epsilon_{\mathrm{noise}}{=}1.0$.


\subsubsection*{\textbf{4) Joint-action Generation}} 
Finally, the joint action was generated by combining the robot's intended action and the AI tutor's predicted action according to the intervention weight. For each proprioceptive dimension, the joint action was defined as
\begin{equation}
    \bar{\bm{x}}^{\mathrm{joint}}_{t+1, i} = (1 - \tilde{\bm{w}}_{t+1,i})\bar{\bm{x}}^{\mathrm{robot}}_{t+1, i} + \tilde{\bm{w}}_{t+1,i}\bar{\bm{x}}^{\mathrm{tutor}}_{t+1, i} 
\end{equation}
where $\bar{\bm{x}}^{\mathrm{joint}}_{t+1,i}$ denotes the $i$-th proprioceptive component of the generated joint action. Smaller values of $\tilde{w}_{t+1,i}$ keep the joint action closer to the robot's intended action, whereas larger values correspond to stronger tutor intervention, pulling the joint action toward the AI tutor's prediction.

Fig.~\ref{fig:ai_tutor} provides an intuitive illustration of how the proposed intervention mechanism operates through online inference. The figure contrasts two representative cases: a low-variability case, in which the tutor expects a narrow range of possible movements, and a high-variability case, in which the tutor expects a broader range of possible movements. These examples illustrate that the same robot--tutor action deviation can result in different interaction dynamics depending on the tutor's expected movement variability.

We next examined whether the advantage of bidirectional tutoring observed in the human-tutoring experiment also emerges under this controlled AI-tutoring framework. To this end, we conducted the AI-tutoring experiment using the same overall experimental design as in the human-tutoring experiment and analyzed both task performance and tutoring dynamics.

\begin{figure}[t]
    \centering
    \includegraphics[width=0.5\textwidth]{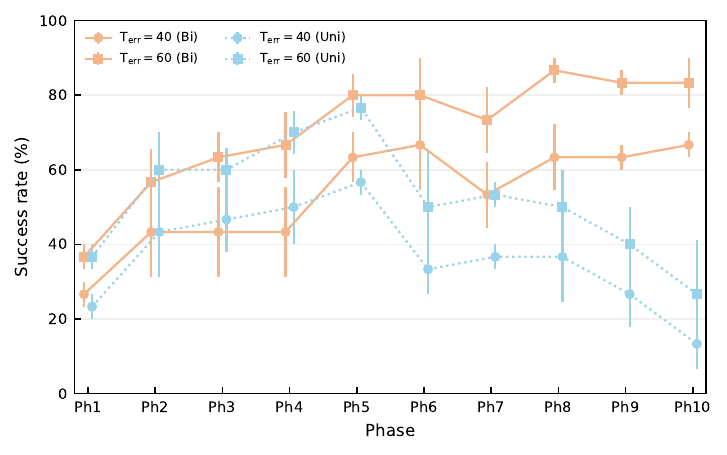}
    \vspace{-8mm}
    \caption{\textbf{Success rate across developmental phases in the AI tutoring experiment.} Success rates are shown for the unidirectional (Uni) and bidirectional (Bi) tutoring conditions. Data points represent the mean success rate across training sets A-C, and error bars indicate the standard error of the mean (SEM). Two spatial error thresholds are used, $T_{\mathrm{err}}{=}40$ and $60$.
    }
    \vspace{-3mm}
    \label{fig:ai_success_rate}
\end{figure}

\subsection{Task Performance}
\begin{figure}[t]
    \centering
    \includegraphics[width=0.51\textwidth]{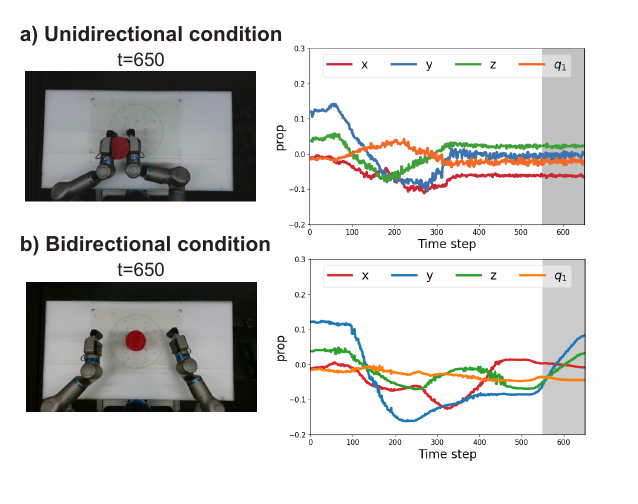}
    \vspace{-13mm}
    \caption{\textbf{Representative testing episode in the AI-tutoring experiment at Phase~10.} Snapshots and corresponding proprioceptive trajectories are shown for the unidirectional and bidirectional conditions. The snapshots show the final state of the episode at $t{=}650$. The gray region indicates the regression window. 
    In this representative episode, the unidirectional condition failed to complete the task: action generation remained near the same posture from around $t{=}300$ until the end of the episode, as reflected in the snapshot. In contrast, the bidirectional condition completed the task successfully, with the snapshot showing the robot returning toward the initial posture after placing the object at the designated return location.    
    }
    \vspace{-3mm}
    \label{fig:ai_test_episode}
\end{figure}

We first evaluated task performance in the AI-tutoring experiment using the same success rate measure as in the human-tutoring experiment. Fig.~\ref{fig:ai_success_rate} shows the success rate across developmental phases for the unidirectional and bidirectional conditions. Overall, the success rate increased during the early developmental phases in both conditions. However, subsequent performance trends differed between the two tutoring conditions. In the bidirectional condition, the success rate continued to increase and remained high in the later phases under both error thresholds. By contrast, in the unidirectional condition, the success rate improved up to the middle phases but declined thereafter. By the final phase, the mean success rate in the bidirectional condition reached $83.3\%$ under $T_{\mathrm{err}}{=}60$ and $66.7\%$ under $T_{\mathrm{err}}{=}40$. In contrast, the corresponding success rates in the unidirectional condition were $26.7\%$ and $13.3\%$, respectively. These results indicate that the advantage of bidirectional tutoring observed in the human-tutoring experiment was also evident under the controlled AI-tutoring framework. Specifically, bidirectional tutoring shows higher and more stable task performance, whereas unidirectional tutoring shows limited improvement followed by a decline in later phases. 
A representative autonomous testing episode is shown in Fig.~\ref{fig:ai_test_episode}. In this example, the unidirectional condition failed to complete the task, whereas the bidirectional condition succeeded. This failure may reflect a mode-averaging-like effect, in which inconsistencies among accumulated tutoring experiences may lead to a poorly organized sensorimotor representation that no longer supports successful task execution. Additional examples of testing episodes are shown in \href{https://youtu.be/8-xmpmk2JKU}{Supplementary Video~3}.

\subsection{Qualitative and Quantitative Comparison in AI-Tutoring Experiment}
\begin{figure*}[tbhp]
    \centering
    \includegraphics[width=0.99\textwidth]{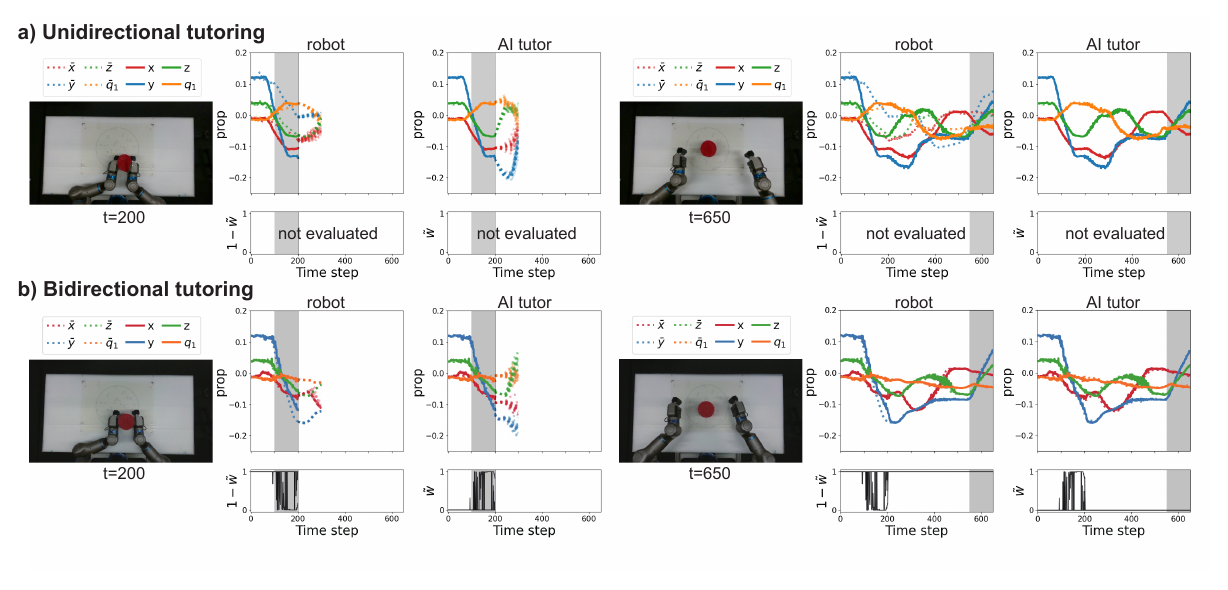}
    \vspace{-8mm}
    \caption{\textbf{Representative examples of AI-tutoring dynamics at Phase~10.} 
    \textbf{a} Unidirectional conditions. \textbf{b} Bidirectional conditions. 
    The examples were selected from the same object-position set. 
    Snapshots are shown at two representative time points: during reaching and grasping ($t{=}200$) and at the end of the episode ($t{=}650$). In each snapshot, the left and right panels show the trajectories evaluated from the robot model and the AI tutor model, respectively. Solid lines indicate the trajectory observed by each model, corresponding to the tutor-imposed trajectory in the unidirectional condition and the co-developed trajectory in the bidirectional condition. Dotted lines indicate each model's reconstructed trajectory before the current time point, and translucent dotted lines after the current time point indicate multiple future predictions generated from prior-sampled latent states. The intervention weight $\tilde{w}$ is shown only in the bidirectional condition and indicates the relative contribution of tutor intervention.
    }
    \label{fig:ai_example_pred_obs}
\end{figure*}

We next examined qualitative and quantitative differences between the unidirectional and bidirectional AI-tutored trajectories, using the same analysis procedure as in the human-tutoring experiment. To provide an intuitive view of the AI-tutoring dynamics, Fig.~\ref{fig:ai_example_pred_obs} shows representative interaction episodes. In the unidirectional condition, the AI tutor generated the tutoring trajectory and imposed it on the robot. As shown at $t{=}650$, the robot-model panel shows a mismatch between the robot model's reconstruction and this imposed observation, whereas the AI-tutor-model panel shows alignment because the observation originated from the tutor model itself. This indicates that the tutor's guidance was externally imposed, rather than being shaped by the robot model's ongoing dynamics. In the bidirectional condition, by contrast, the observed trajectory was co-developed through interaction between the robot and the AI tutor. As shown at $t{=}650$, the robot-model panel shows that the co-developed observation remained aligned with the robot model's reconstruction, suggesting that the resulting trajectory became relatively consistent with the robot's ongoing dynamics. Importantly, this alignment did not result from the robot acting alone. During the reaching period toward the newly introduced object location ($t{=}100$--$200$), the increased intervention weight indicated that the AI tutor intervened when the robot model's prediction was not yet well adapted to the new object position. At the same time, the tutor-model panel shows that the tutor's reconstruction also remained aligned with the co-developed observation even when the intervention weight was near zero. This suggests that the tutor updated its internal state in accordance with the robot's ongoing behavior, allowing its guidance to remain closely coupled with the robot model's ongoing dynamics rather than being externally imposed. The corresponding video of this episode and additional examples are provided in \href{https://youtu.be/5mu1MfNB24k}{Supplementary Video~4}.

\subsubsection*{\textbf{Spatiotemporal Structure of AI-Tutored Trajectories}}
\begin{figure*}[tbhp]
    \centering
    \includegraphics[width=0.99\textwidth]{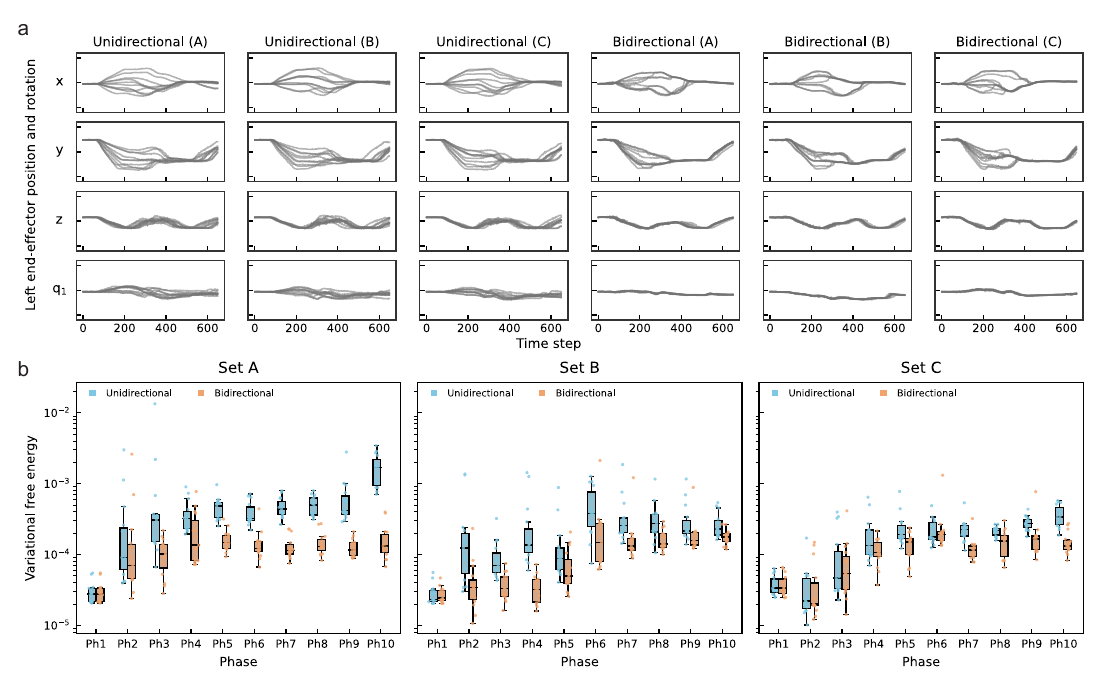}
    \vspace{-3mm}
    \caption{\textbf{Qualitative and quantitative comparison of AI-tutoring dynamics under the unidirectional and bidirectional conditions.} 
    \textbf{a} Left end-effector position and rotation trajectories obtained during AI-tutoring sessions, overlaid across the ten developmental phases for each training set (A--C). Each trace corresponds to the tutored trajectory at one phase. For rotation, only the first quaternion component ($q_1$) is shown, with the other components ($q_2$, $q_3$, and $q_4$) omitted for clarity. 
    \textbf{b} Training loss, quantified as variational free energy, across developmental phases in the AI-tutoring experiment. Box plots show the distribution across random seeds. Statistical analysis was conducted on log-transformed values from Phases 2--10. Phase 1 was excluded from the statistical analysis because it served as a common baseline before the tutoring conditions diverged.
    }
    \label{fig:ai_tutoring_dynamics}
\end{figure*}

Fig.~\ref{fig:ai_tutoring_dynamics}-a shows the left end-effector position and rotation trajectories obtained during AI tutoring in the unidirectional and bidirectional conditions. Consistent with the human-tutoring experiment, the unidirectional condition showed larger spatiotemporal variability across developmental phases, whereas the bidirectional condition produced more organized trajectories. In the bidirectional condition, task-critical variability remained around the reaching period, reflecting adjustment to newly introduced object locations, but overall trajectories were more aligned in shape and timing than in the unidirectional condition. Compared with the human-tutoring experiment, however, the bidirectional AI-tutored trajectories appeared to retain somewhat larger variability. This may reflect the more limited adaptability of the AI tutor, whose intervention was constrained by the learned tutor model and predefined intervention mechanism. 
Nevertheless, the key trajectory-level pattern also emerged under the AI-tutoring framework: bidirectional tutoring produced more organized trajectories than unidirectional tutoring while still allowing task-critical adjustments for newly introduced object locations.

\subsubsection*{\textbf{Training Loss and Trajectory Compatibility in AI-Tutoring Experiment}}
We next examined the training loss, quantified as the variational free energy. As in the human-tutoring experiment, the loss value was interpreted as an index of how compatible the tutored trajectories were with the model's previously acquired sensorimotor dynamics. 


As shown in Fig.~\ref{fig:ai_tutoring_dynamics}-b, the bidirectional condition generally yielded lower training loss than the unidirectional condition across developmental phases and training sets. This difference remained significant in a linear model that accounted for developmental phase and training set (condition effect for bidirectional tutoring relative to unidirectional tutoring: $\beta{=}{-}0.805$, SE${=}0.052$, $t{=}{-}15.50$, $p{<}0.001$). Thus, the quantitative training-loss pattern observed in the human-tutoring experiment was also found under the AI-tutoring framework. However, the separation between conditions was smaller than that observed in the human-tutoring experiment. 
This smaller difference may be related to the trajectory-level pattern shown in Fig.~\ref{fig:ai_tutoring_dynamics}-a. Specifically, the bidirectional AI-tutored trajectories retained greater variability than those observed in the human-tutoring experiment, possibly reflecting the limited adaptability of the AI tutor compared with the human tutor. Nevertheless, the bidirectional condition still yielded lower training loss than the unidirectional condition, indicating that the same overall pattern was observed under the controlled AI-tutoring framework.


\subsubsection*{\textbf{Reduction of AI Tutor Intervention across Developmental Phases}}
\begin{figure}[t]
    \centering
    \includegraphics[width=0.48\textwidth]{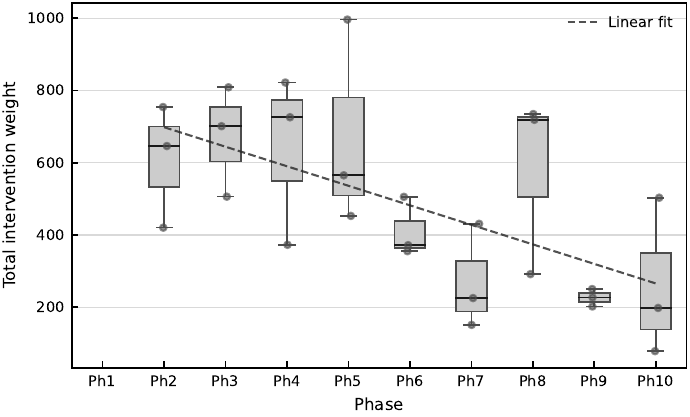}
    \vspace{-2mm}
    \caption{\textbf{Changes in AI tutor intervention during bidirectional tutoring.} Box plots show the total intervention weight applied as tutor intervention during each tutoring episode in the bidirectional condition. Data from training sets A–C were pooled within each developmental phase. The dashed line indicates the linear trend estimated by linear regression across developmental phases. Total intervention weight decreased over phases (Pearson's $r{=}{-}0.781$, $p{=}0.0130$; linear regression $R^{2}{=}0.610$). No value is shown for Ph1 because this baseline phase did not include robot-to-tutor physical feedback, which was required to compute this measure.
    }
    \label{fig:ai_intervention_weight}
    \vspace{-2mm}
\end{figure}

Finally, we examined the amount of AI tutor intervention in the bidirectional condition as an additional behavioral index of learning dynamics. Here, tutor intervention was quantified by the total intervention weight accumulated over each tutoring episode, $\sum_{t=1}^{T} \tilde{\bm{w}}_{t}$. As shown in Fig.~\ref{fig:ai_intervention_weight}, the total intervention weight showed an overall decreasing trend across developmental phases. As in the human-tutoring experiment, this reduction suggests that the robot gradually required less tutor intervention across developmental phases. Together with the increased task success rate observed in Fig.~\ref{fig:ai_success_rate}, this result indicates that the robot acquired task-relevant motor skills while becoming less dependent on AI tutor guidance.

\section{Discussion}
In this study, we hypothesized that bidirectional tutoring supports stable developmental motor learning through the formation of co-developed trajectories that preserve behavioral coherence. The present results are consistent with this hypothesis across both human- and AI-tutoring experiments. In both settings, bidirectional tutoring resulted in higher and more stable task performance than unidirectional tutoring, particularly in later developmental phases (Fig.~\ref{fig:human_sr}, Fig.~\ref{fig:ai_success_rate}).

Previous studies in robot learning from demonstration have emphasized that the quality and consistency of tutoring trajectories are crucial to forming stable sensorimotor representations and supporting generalization~\cite{tani2003self, wang2025generalization, sakr2025consistency}. The present results support and extend this view by showing that such consistency can arise in a developmental tutoring process. In both human- and AI-tutoring experiments, bidirectional tutoring produced more spatiotemporally coherent trajectories than those in the unidirectional condition (Fig.~\ref{fig:human_traj_loss}-a, Fig.~\ref{fig:ai_tutoring_dynamics}-a). 
These trajectories appeared to be consistent with the robot's previously acquired behavioral patterns. This suggests that, in bidirectional tutoring, the resulting tutoring experiences were naturally constrained by the robot's current generative dynamics, while tutor intervention introduced corrections for newly encountered task-relevant variations. Through this coupling, the co-developed trajectories maintained behavioral coherence across phases.
The reduction of tutor intervention across developmental phases further supports the view that the coherent trajectories were co-developed rather than externally imposed by fixed tutor guidance (Fig.~\ref{fig:human_torque}, Fig.~\ref{fig:ai_intervention_weight}). This indicates that the tutor did not simply ignore the robot's developing behavior, but adjusted its intervention according to the robot's improving motor skills.

This view is closely related to the broader idea that learning experiences are not fixed inputs, but are shaped by the learner’s own behavior during ongoing interaction. For example, previous studies in reinforcement learning and imitation learning have discussed the influence of learner behavior on subsequent learning experiences in closed-loop interactions with the environment~\cite{sutton1998reinforcement, liu2019off, ross2011reduction}. A related perspective is also found in sensorimotor contingency theory, which emphasizes the closed-loop coupling between action and sensory change~\cite{o2001sensorimotor, lubbert2021socializing, jacquey2019sensorimotor}. The present study extends this closed-loop perspective to a tutoring context~\cite{wood1976role}, \ie socially grounded context in which learning experiences are co-developed through bidirectional interaction between the tutor and the learner. Specifically, bidirectional tutoring naturally incorporates the learner's internal dynamics, allowing them to guide the formation of consistent behavioral patterns.




Several limitations should be noted. First, the AI tutor reproduced only selected functional aspects of human tutoring, not the full richness of human adaptive behavior. However, it provided a controlled interaction setting under a fixed tutoring policy, which was sufficient for testing whether such bidirectional coupling could shape the resulting trajectories and support stable learning in the present study. Second, the present study was limited to a simple object manipulation task. In addition, the human-tutoring experiment involved a single tutor, although the procedure was repeated across three independently sampled object-position sets. These restrictions were partly due to the experimental cost of stage-wise developmental learning: even in this simple task, each experiment required repeated physical tutoring, model retraining, and evaluation across developmental phases. Thus, systematic comparisons involving multiple human participants, more complex tasks would require substantially more experimental time. Future work should address both the limited adaptivity of the AI tutor and the experimental constraints of the present study by developing more autonomous and adaptive AI tutor agents. For example, large language models could support high-level interpretation of the robot's ongoing behavior and the generation of richer and more adaptive tutoring policies across developmental phases. Such agents could enable larger-scale and more systematic tutoring experiments, thereby reducing the burden of human-participant studies while allowing the resulting interaction principles to be validated later in real embodied human--robot interaction.

In summary, these findings highlight bidirectional tutoring as a co-developmental process in which new learning experiences are shaped jointly by tutor guidance and the learner's prior experience, rather than as the transfer of externally imposed demonstrations. This perspective may support the extension of bidirectional tutoring as an embodied and socially grounded approach to developmental motor learning in robots.
\section*{Acknowledgments}
J.T. was partially funded by the Japan Society for the Promotion of Science (JSPS) KAKENHI, Transformative Research Area (A): unified theory of prediction and action [26H01186].

\appendix
\subsubsection*{\textbf{PV-RNN Forward Computation}}\label{app:network computation}
Its computation of the hidden state $\bm{d}_t$ and prediction $\bar{\bm{x}}_t$ are as follows.
\begin{align} 
    \bm{h}_{t} &= (1-\frac{1}{\tau})\bm{h}_{t-1} + \frac{1}{\tau}(\bm{W}_{hd}\bm{d}_{t-1}+\bm{W}_{hz}\bm{z}_{t}+\bm{b}_{h})\label{eq:ctrnn_h}\\
    \bm{d}_{t} &= \mathrm{tanh}{(\bm{h}_{t})}\label{eq:ctrnn_d}\\
    \bar{\bm{x}}_{t} &= \mathrm{tanh}{(\bm{W}_{\mathrm{out}}\bm{d}_{t} + \bm{b}_{\mathrm{out}})}\label{eq:ctrnn_out}
\end{align}
where $\bm{h}_t$ represents the hidden state before applying an activation function. 
$\tau$ is the time constant that determines how strongly past states contribute to the current state~\cite{beer1995dynamics}. $\bm{W}_{hd}$, $\bm{W}_{hz}$, and $\bm{W}_{\mathrm{out}}$ are weight matrices. $\bm{b}_{h}$ and $\bm{b}_{\mathrm{out}}$ are bias terms. $\bm{z}_{t}$ is depended on prior $\bm{z}^{p}_{t}$ or posterior $\bm{z}^{q}_{t}$. $\bm{z}^{p}_{t}$ is represented as a distribution with mean $\bm{\mu}^{p}_{t}$ and variance $\bm{\sigma}^{p}_{t}$ as follows.
\begin{align} \label{eq:mu_dtod}
\bm{\mu}^{p}_{t} &= \tanh{(\bm{W}_{\mu}\bm{d}_{t-1}+\bm{b}_{\mu})} \\
\bm{\sigma}^{p}_{t} &= \text{sigmoid}(\bm{W}_{\sigma}\bm{d}_{t-1}+\bm{b}_{\sigma})\\
\bm{z}^{p}_{t} &= \bm{\mu}^{p}_{t} + \bm{\sigma}^{p}_{t}*\bm{\epsilon}_{t}
\end{align} 
where $\bm{\epsilon}_{t}$ is noise sampled from a standard normal distribution $\mathcal{N}(0, 1)$~\cite{kingma2013auto}. Posterior $\bm{z}^{q}_t$ is computed as follows
\begin{align} 
\bm{\mu}^{q}_{t} &= \tanh{(\bm{a}^{\mu}_{t})}\\
\bm{\sigma}^{q}_{t} &= \text{sigmoid}(\bm{a}^{\sigma}_{t})\\
\bm{z}^{q}_{t} &= \bm{\mu}^{q}_{t} + \bm{\sigma}^{q}_{t}*\bm{\epsilon}_{t}
\end{align} 

In our model, maximizing the ELBO corresponds to the maximizing expected log-likelihood of the observation given the model’s prediction (accuracy term: $e_{t}$). Assuming a Gaussian likelihood with unit variance, the conditional probability can be expressed as
\begin{align} 
\ln{p(\bm{x}_{t}|\bm{d}_{t})}  \propto -\frac{1}{2}\left||\bm{x}_{t}-\bar{\bm{x}}_{t}\right||^{2}
\end{align} 
and thus maximizing the log-likelihood is equivalent to minimizing the squared reconstruction error, defined as:
\begin{align} 
e_{t} &= \frac{1}{2N_{x}}\sum^{N_x}_{i=1}||\bm{x}_{t, i}-\bar{\bm{x}}_{t, i}||^2
\end{align} 
where $N_x$ is the dimension of $\bm{x}$. $i$ represents the $i$th entry of each vector. Conversely, maximizing the ELBO also corresponds to minimizing the Kullback–Leibler divergence (KLD) between the approximate posterior and the prior (complexity term: $r_{t}$), defined as:
\begin{align}\label{eq:KLD loss computation}
    r_{t} 
    &= \frac{1}{N_{z}}D_{\mathrm{KL}}[q_{\phi}(\bm{z}_{t}|\bm{d}_{t-1}, \bm{e}_{t:T})||p_{\theta}(\bm{z}_{t}|\bm{d}_{t-1})]\\
    &=\frac{1}{N_{z}}\sum^{N_z}_{i=1}\left\{ \ln{\frac{\bm{\sigma}^{p}_{t,i}}{\bm{\sigma}^{q}_{t,i}}} + \frac{(\bm{\mu}^{q}_{t,i}-\bm{\mu}^{p}_{t,i})^{2}+(\bm{\sigma}^{q}_{t,i})^{2}}{2(\bm{\sigma}^{p}_{t,i})^{2}} - \frac{1}{2}\right\} 
\end{align} 
where $N_{z}$ is the dimension of $\bm{z}$. $i$ represents the $i$th entry of each vector.

\subsubsection*{\textbf{Detailed Procedure for Online Inference and Action Generation}}\label{app:network computation er}

\begin{algorithm}[t]
    \caption{Error regression and action generation}
    \label{alg:er}
    \begin{algorithmic}[1]
        \For{$t = 1$ to $T$}
            \State Initialize the adaptive vectors at time $t$:
            \[
            \bm{a}^{\mu}_{t}\leftarrow \mathrm{arctanh}(\bm{\mu}^{p}_{t}),\quad
            \bm{a}^{\sigma}_{t}\leftarrow \mathrm{logit}(\bm{\sigma}^{p}_{t})
            \]            
            \State Observe new targets $\bm{x}_t$ 
            \State Growing and shifting window  
            \[
            t_{\mathrm{start}} \gets \max(1, t-t_{\mathrm{win}})
            \]
            \For{$i=1$ to $N_{\mathrm{itr}}$} 
                \State Generate $\bar{\bm{x}}_{t_{\mathrm{start}:t}}$ with $\bm{a}_{t_{\mathrm{start}}:t}$ 
                \State Compute $\mathcal{L}(\bm{\phi})$ and BPTT between $t_{\mathrm{start}}$ and $t$ 
                \State Update $\bm{a}_{t_{\mathrm{start}}:t}$ 
            \EndFor
            \State Generate $\bar{\bm{x}}_{t_{\mathrm{start}:t}}$ with $\bm{a}_{t_{\mathrm{start}}:t}$ 
            \State Generate prediction $\bar{\bm{x}}_{t+1}$ with sampling from the prior 
            \State Send $\bar{\bm{x}}_{t+1,\mathrm{prop}}$ to the controller for action generation
        \EndFor
    \end{algorithmic}
\end{algorithm}

Algorithm~\ref{alg:er} shows the detailed procedure for online inference and action generation through error regression with a growing and shifting window. As the time step advances, the adaptive vectors for the approximate posterior at the new time step are initialized from the corresponding prior using the inverse transformations of those used to compute the posterior parameters, namely $\mathrm{arctanh}(\cdot)$ and $\mathrm{logit(\cdot)}$ (see Appendix~\ref{app:network computation}, Forward Computation). The next-step prediction is then generated by sampling from the prior, and its proprioceptive component is provided to the controller for action generation.

\bibliographystyle{IEEEtran}
\bibliography{bib}

\newpage
\section*{Supplementary Material}
\subsection{Generation of AI-Tutor Training Trajectories}
\label{supp:ai_tutor_training_trajectory_generation}

The AI-tutor training trajectories were generated computationally rather than recorded from human demonstrations. The purpose of this procedure was to provide task-successful trajectories that preserved the overall manipulation sequence while including spatiotemporal variability in detailed movement execution. The reference object-grasping postures of both end effectors were defined based on a successful center-object grasping configuration. Each generated trajectory followed the same task-level sequence of reaching, grasping, lifting, placing, releasing, and returning. The sequence was specified by task-relevant waypoints,
\[
\begin{aligned}
\mathcal{V}=\{&
v_{\mathrm{home}},
v_{\mathrm{reach}},
v_{\mathrm{grasp}},
v_{\mathrm{lift}},\\
&
v_{\mathrm{above\ center}},
v_{\mathrm{place\ center}},
v_{\mathrm{return}}
\}.
\end{aligned}
\]
which corresponded to the object-reaching position, lift-up position, position above the center of the workspace, placing position at the center of the workspace, and return position near the arm home posture, respectively. Here, \(v_{\mathrm{grasp}}\) was defined by shifting the left and right end-effector components of \(v_{\mathrm{reach}}\) by 1 cm toward the object center along the lateral direction, so that the hands moved inward to grasp the object.

Variability was introduced by modifying the task-relevant waypoints in \(\mathcal{V}\) in a structured manner. First, object-location variability was introduced by translating the reaching waypoint \(v_{\mathrm{reach}}\) according to one of 37 object-location conditions defined on a grid over the workspace. The same 2D displacement was applied to the left and right end-effector components of \(v_{\mathrm{reach}}\), thereby changing the target object location while preserving the relative bimanual grasping configuration. Second, movement-shape variability was introduced by perturbing selected non-critical waypoints in \(\mathcal{V}\). Specifically, random Cartesian perturbations were applied to \(v_{\mathrm{lift}}\), \(v_{\mathrm{above\ center}}\), and \(v_{\mathrm{return}}\), while \(v_{\mathrm{place\ center}}\) was kept fixed to preserve the designated return location. For \(v_{\mathrm{lift}}\) and \(v_{\mathrm{above\ center}}\), the same Cartesian perturbation was applied to both end effectors, sampled uniformly within \(\pm 5\) cm along each Cartesian axis. For \(v_{\mathrm{return}}\), independent Cartesian perturbations were applied to the left and right end effectors, also sampled uniformly within \(\pm 5\) cm. This allowed the generated trajectories to vary in detailed movement shape while preserving task-critical reaching and placing constraints. Finally, rotational variability was introduced by perturbing the end-effector orientations associated with the selected waypoints. Random Euler-angle perturbations were sampled uniformly within \(\pm 5^\circ\) around the \(x\) and \(y\) axes and \(\pm 10^\circ\) around the \(z\) axis, and the resulting orientations were converted to quaternion targets.

Temporal variability was introduced by randomly perturbing the duration of each movement segment. The base segment durations are shown in Table~\ref{tab:supp_ai_tutor_durations}. For each generated sequence, the duration of each segment except the final one was randomly jittered within approximately one third of its base duration, and the final segment duration was adjusted so that the total sequence length remained fixed at $T{=}650$. This changed the timing of reaching, grasping, lifting, placing, and returning while preserving the overall task sequence

After the waypoints and segment durations were determined, each trajectory was generated by interpolating between successive waypoints for both position and orientation. The generation procedure was repeated for 37 object-location conditions and 15 random seeds, resulting in 555 AI-tutor training trajectories. Because the same task-level waypoint structure was used across all trajectories, the generated sequences preserved the overall manipulation sequence. At the same time, they differed in object location, movement timing, movement shape, and end-effector orientation. Thus, the AI tutor was trained on variable task-successful executions rather than repeated deterministic templates.

\begin{table}[t]
    \centering
    \caption{Base segment durations.}
    \label{tab:supp_ai_tutor_durations}
    \footnotesize
    \begin{tabular}{lc}
        \toprule
        Segment (waypoint relation) & Steps \\
        \midrule
        Home (stay at \(v_{\mathrm{home}}\)) & 75 \\
        Reach (\(v_{\mathrm{home}}\!\rightarrow\!v_{\mathrm{reach}}\)) & 75 \\
        Grasp (\(v_{\mathrm{reach}}\!\rightarrow\!v_{\mathrm{grasp}}\)) & 10 \\
        Stay after grasping (stay at \(v_{\mathrm{grasp}}\)) & 30 \\
        Lift (\(v_{\mathrm{grasp}}\!\rightarrow\!v_{\mathrm{lift}}\)) & 45 \\
        Stay after lifting (stay at \(v_{\mathrm{lift}}\)) & 30 \\
        Move above (\(v_{\mathrm{lift}}\!\rightarrow\!v_{\mathrm{above}}\)) & 45 \\
        Stay above (stay at \(v_{\mathrm{above}}\)) & 30 \\
        Put down (\(v_{\mathrm{above}}\!\rightarrow\!v_{\mathrm{place}}\)) & 45 \\
        Stay after placing (stay at \(v_{\mathrm{place}}\)) & 80 \\
        Release (open hands at \(v_{\mathrm{place}}\)) & 10 \\
        Return (\(v_{\mathrm{place}}\!\rightarrow\!v_{\mathrm{return}}\)) & 75 \\
        Final stay (stay at \(v_{\mathrm{return}}\)) & 100 \\
        \bottomrule
    \end{tabular}
\end{table}

\subsection{Supplementary Videos}
\label{supp:videos}
\begin{itemize}
    \item \href{https://youtu.be/cgVFxeJdcck}{Supplementary Video~1} shows representative autonomous testing episodes in the human-tutoring experiment.
    \item \href{https://youtu.be/GxxSs259urM}{Supplementary Video~2} shows representative human-tutoring episodes.
    \item \href{https://youtu.be/8-xmpmk2JKU}{Supplementary Video~3} shows representative autonomous testing episodes in the AI-tutoring experiment.
    \item \href{https://youtu.be/5mu1MfNB24k}{Supplementary Video~4} shows representative AI-tutoring episodes.
\end{itemize}

\subsection{Data and Code Availability}
The processed data and analysis code used in this study are available at: \url{https://github.com/oist-cnru/tcds-motor-skill-tutoring-code-data}.




\vfill

\end{document}